\documentclass[runningheads]{llncs}

\usepackage{eccv}

\usepackage{eccvabbrv}

\usepackage{graphicx}
\usepackage{booktabs}

\usepackage[accsupp]{axessibility}  
\usepackage{pifont}      
\usepackage{bbding}       
\usepackage{fontawesome}  

\usepackage{hyperref}

\usepackage{orcidlink}
\usepackage{amsmath}
\usepackage{pifont}
\usepackage{bbm}
\usepackage{color}
\usepackage{booktabs}
\usepackage{makecell}
\usepackage{mathrsfs}
\usepackage{multirow} 
\usepackage[ruled, linesnumbered, noend]{algorithm2e}

\begin{document}

\title{Refine, Discriminate and Align:\\Stealing Encoders via Sample-Wise Prototypes and Multi-Relational Extraction} 

\titlerunning{Refine, Discriminate and Align (RDA)}

\author{
Shuchi Wu$^{1}$ ~
Chuan Ma$^{2}$\textsuperscript{\Envelope} ~
Kang Wei$^{3}$\textsuperscript{\Envelope} ~
Xiaogang Xu$^{4}$ ~
Ming Ding$^{5}$ ~ 
\\Yuwen Qian$^{1}$ ~
Di Xiao$^{2}$ ~
Tao Xiang$^{2}$
\\
$^1$ NJUST \quad
$^2$ CQU  \quad 
$^3$ PolyU \quad 
$^4$ CUHK \quad
$^5$ Data61, CSIRO
}

\authorrunning{S. Wu et al.}

\institute{}

\maketitle \let\thefootnote\relax\footnotetext{\textsuperscript{\Envelope} Corresponding Author: \email{chuan.ma@cqu.edu.cn, adam-kang.wei@polyu.edu.hk}}

\begin{abstract}
    This paper introduces \textbf{RDA}, a pioneering approach designed to address two primary deficiencies prevalent in previous endeavors aiming at stealing pre-trained encoders: (1) suboptimal performances attributed to biased optimization objectives, and (2) elevated query costs stemming from the end-to-end paradigm that necessitates querying the target encoder every epoch. Specifically, we initially \textbf{\underline{R}}efine the representations of the target encoder for each training sample, thereby establishing a less biased optimization objective before the steal-training phase. This is accomplished via a sample-wise prototype, which consolidates the target encoder's representations for a given sample's various perspectives. Demanding exponentially fewer queries compared to the end-to-end approach, prototypes can be instantiated to guide subsequent query-free training. For more potent efficacy, we develop a multi-relational extraction loss that trains the surrogate encoder to \textbf{\underline{D}}iscriminate mismatched embedding-prototype pairs while \textbf{\underline{A}}ligning those matched ones in terms of both amplitude and angle. In this way, the trained surrogate encoder achieves state-of-the-art results across the board in various downstream datasets with limited queries. Moreover, RDA is shown to be robust to multiple widely-used defenses. Our code is available at \url{https://github.com/ShuchiWu/RDA}.
\end{abstract}

\begin{figure}[t]
    \centering
    \includegraphics[width=0.8\textwidth]{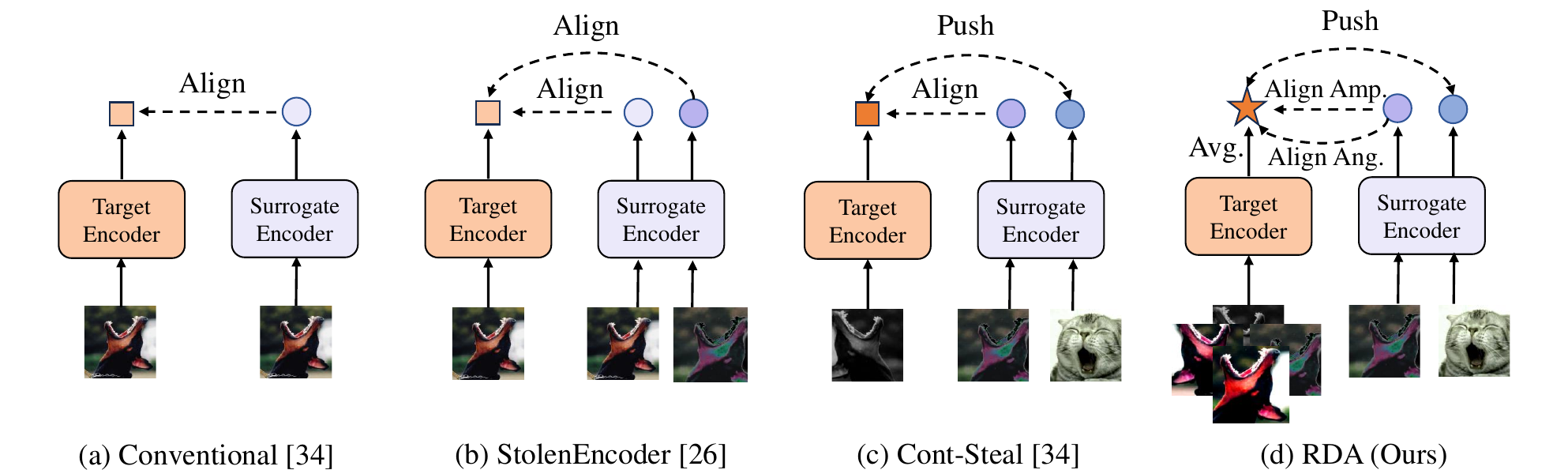}
    \caption{Illustrations of four stealing methods against SSL. The dotted arrows and text beside interpret how each method optimizes the surrogate encoder. Surrogate encoder branches in (b)-(d) involve data augmentations for training. Both (c) and (d) augment each sample before querying the target encoder but adopting different schemes.}
    \label{fig: existing methods}
\end{figure}

\section{Introduction}
\label{sec:intro}
Self-supervised learning (SSL) \cite{devlin2018bert, chen2020simple, he2020momentum, grill2020bootstrap, caron2021emerging} is endowed with the capability to harness unlabeled data for pre-training a versatile encoder that applicable to a range of downstream tasks, or even showcasing groundbreaking zero-shot performances, e.g., CLIP \cite{radford2021learning}. However, SSL typically requires a large volume of data and computation resources to achieve a convincing performance, i.e., high-performance encoders are expensive to train \cite{sharir2020cost}. 
To safeguard the confidentiality and economic worth of these encoders, entities like OpenAI offer their encoders as a premium service, exclusively unveiling the service API to the public. Users have the privilege of soliciting embeddings for their data, facilitating the training of diverse models tailored for specific downstream tasks. Regrettably, the substantial value of these encoders and their exposure to publicly accessible APIs render them susceptible to model stealing attacks \cite{liu2022stolenencoder, sha2023can, dziedzic2022difficulty}.

Given a surrogate dataset, model stealing attacks aim to mimic the outputs of a target model to train either a high-accuracy copy of comparable performance or a high-fidelity copy that can serve as a stepping stone to perform further attacks like adversarial examples \cite{papernot2017practical}, membership inference attacks \cite{shokri2017membership,salem2018ml}, etc. The goal of this paper is to develop an approach that can train a surrogate encoder with competitive performance on downstream tasks by only accessing the target encoder's output embeddings. Meanwhile, it should require as little cost as possible, which is primarily derived from querying the target encoder.

\begin{figure}[t]
    \centering
    \includegraphics[width=0.35\textwidth]{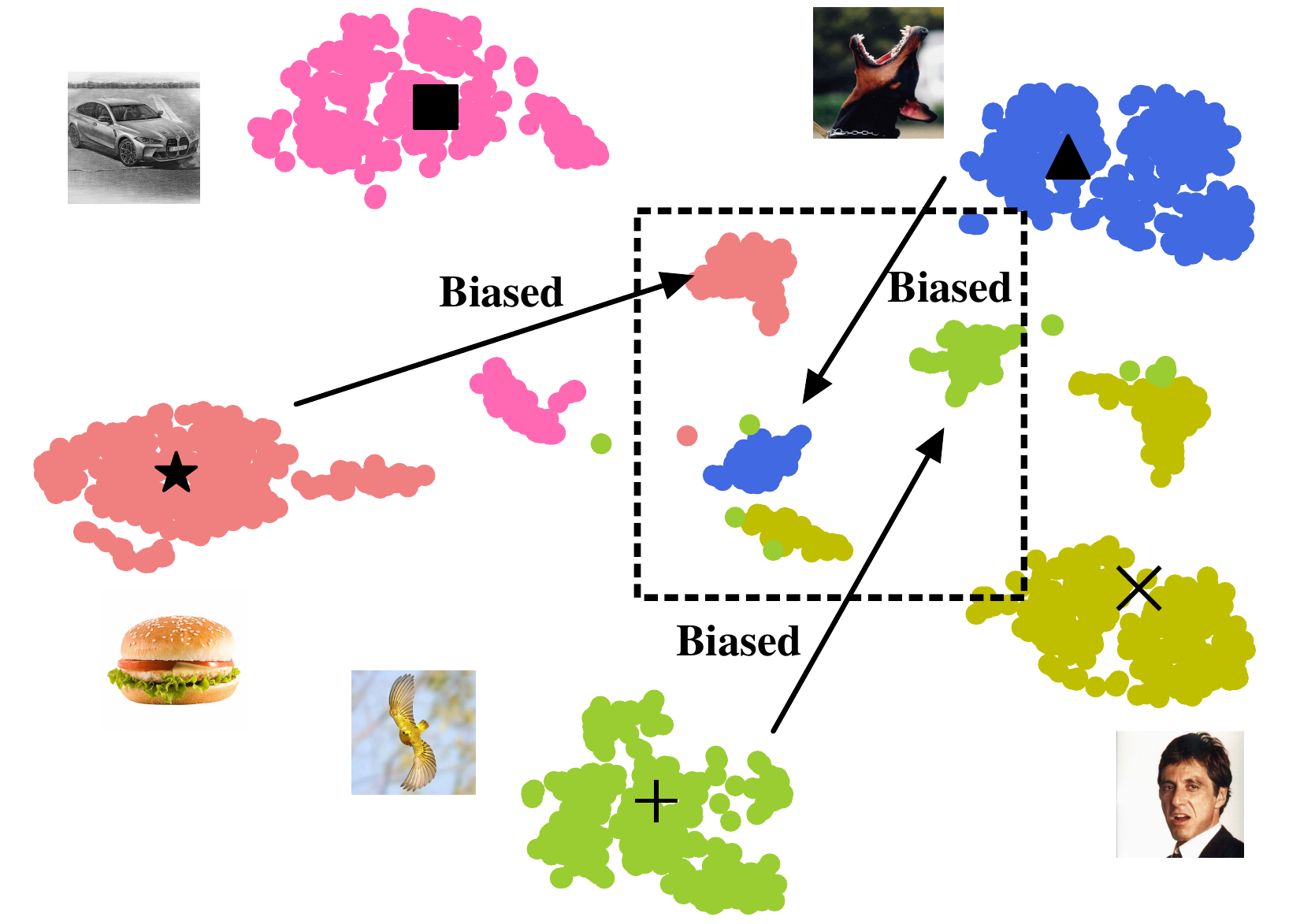}
    \caption{t-SNE of embeddings belonging to five different images generated by an encoder pre-trained on CIFAR10, with each image augmented into 500 patches and fed into the encoder. \textbf{Each black marker represents the mean of the 500 embeddings of a certain image, i.e., its prototype.} Among the embeddings of an image's various augmentations, some can be \textbf{diverged} or even \textbf{biased}. In contrast, each image's prototype is \textbf{more distinguishable}, i.e., \textbf{less biased}.}
    \label{fig:motivation}
\end{figure}

We begin with systematically studying existing techniques for stealing pre-trained encoders. Specifically, the conventional method \cite{sha2023can} and StolenEncoder \cite{liu2022stolenencoder} are similar, and both only need to query the target encoder once with each sample before the training. The output embedding can be viewed as a ``ground truth'' or an ``optimization objective'' to the sample, and the surrogate encoder is optimized to output a similar embedding when the same sample is fed, as illustrated in Figure \ref{fig: existing methods} (a). The sole distinction lies in the optimization of StolenEncoder, which incorporates data augmentations (refer to Figure \ref{fig: existing methods} (b)), as it supposes the target encoder will produce similar embeddings for an image and its augmentations. Nevertheless, we contend that this presumed similarity is of a modest degree, given the disparity between the surrogate and pre-training data. The target encoder's representations for such data's various augmentations may diverge and even be biased to other samples, as visually demonstrated in Figure \ref{fig:motivation} using a toy experiment. In this sense, only using the embedding of an image's single perspective (regardless of the original or augmented version) as the optimization objective is inadvisable.

On the other hand, Cont-Steal \cite{sha2023can} augments each sample into two perspectives in each epoch: one is used to query the target encoder while the other is fed to the surrogate encoder. Two embeddings of the same sample from the target and surrogate encoders will be aligned, while those of different samples will be pushed apart, as illustrated in Figure \ref{fig: existing methods} (c). In spite of such an end-to-end training scheme performing better, it suffers from high query costs since each sample is augmented and used to query in each epoch.

To tackle these issues, we propose \textbf{\textit{RDA}}. Specifically, we first \textbf{\underline{R}}efine the target encoder's representations for each sample in the surrogate dataset by averagely aggregating its several (e.g., 10) different augmentations' embeddings, i.e., their mean value, to establish a sample-level prototype for it. Then, the sample-wise prototype is used to guide the surrogate encoder optimization, which can mitigate the impact of biased embeddings, as shown in Figure \ref{fig:motivation} (see black markers). We have an experiment presented in Figure 10 of our supplementary material to further quantitatively reveal the benefit of prototypes, i.e., significantly more similar with each augmentation patch's embedding, showcasing it is less biased. With a prototype, there is a static optimization objective for each sample across the entire training, and thus, it is done in a query-free manner. This enables RDA to have far less query cost than the end-to-end approach (less than 10\%). To further enhance the attack efficacy with limited queries, we develop a \textit{multi-relational extraction loss} that trains the surrogate encoder to \textbf{\underline{D}}iscriminate mismatched embedding-prototype pairs while \textbf{\underline{A}}ligning those matched ones in terms of both amplitude and angle. The framework and pipeline of RDA are depicted in Figures \ref{fig: existing methods} (d) and \ref{fig: pipeline}, respectively.

\begin{figure}[t]
    \centering
    \includegraphics[width=0.52\textwidth]{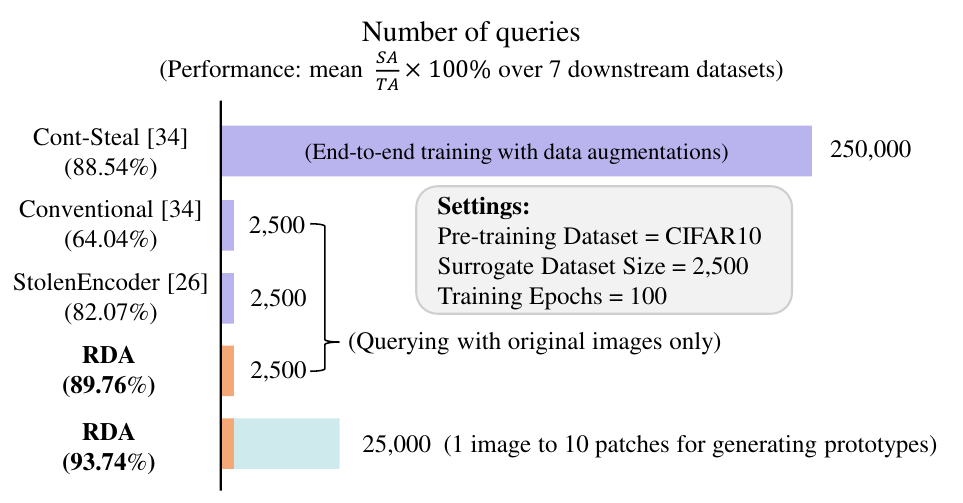}
    \caption{Performance comparisons between four stealing methods against SSL. The presented results are the mean values achieved by each method over seven different downstream classification tasks, with their corresponding query costs. \textbf{Our proposed RDA can achieve SOTA results with the least query cost}.} 
    \label{fig: results summary}
\end{figure}

Extensive experiments show that RDA acquires a favorable trade-off between the query cost (a.k.a, the money cost) and attack efficacy. As depicted in Figure \ref{fig: results summary}, when the target encoder is pre-trained on CIFAR10, RDA averagely outperforms the state-of-the-art (SOTA), i.e., Cont-Steal \cite{sha2023can}, by 1.22\% on the stealing efficacy over seven different downstream datasets, with a competitive smaller query cost, i.e., only 1\% of Cont-Steal. Further, this performance gap can be widened to 5.20\% with 10\% query cost of Cont-Steal. In particular, RDA is demonstrated to be robust against multiple prevalent defenses \cite{orekondy2019knockoff, tramer2016stealing, jia2022badencoder}. 

In conclusion, our contributions are three folds:
\begin{itemize}
    \item We comprehensively investigate two critical inadequacies of existing stealing methods against SSL, i.e., suboptimal efficacy and high query costs, and analyze the causes. 
    \item We develop a novel approach to train surrogate encoders, namely RDA, with sample-wise prototype guidance and a multi-relational extraction loss.
    \item Extensive experiments are conducted to verify the effectiveness and robustness of RDA.
\end{itemize}

\section{Related Work}

\noindent
\textbf{Prototype Learning.}\quad A prototype refers to the mean of embeddings belonging to all images of an identical class, which serves as a proxy of the class \cite{tan2022fedproto}. With each class having a prototype, the model is trained to match its output with the corresponding prototype when inputting a sample of a certain class. Prototype learning has proven beneficial for various learning scenarios, e.g., federated learning \cite{huang2023rethinking, dai2023tackling, tan2022federated} and few-shot learning \cite{mettes2019hyperspherical, snell2017prototypical, tian2020rethinking}.
In this paper, the optimization objective (i.e., the mean of embeddings belonging to an identical sample's multiple augmentations) we generate for each training sample is conceptually similar to prototypes, and thus we name it a sample-wise prototype. 

\noindent
\textbf{Model Stealing Attacks against SSL.}\quad Dziedzic \textit{et al.} \cite{dziedzic2022difficulty} pointed out that the higher dimension of embeddings 
leaks more information than labels, 
making SSL more easily stolen. The primary objective of a stealing attack is to train a surrogate encoder to achieve high accuracy on downstream tasks or recreate a high-fidelity copy that can be used to mount further attacks such as adversarial examples \cite{carlini2017towards} and membership inference attacks \cite{liu2021encodermi}.

\noindent
\textbf{Defenses against Model Stealing Attacks.}\quad Existing defenses against model stealing attacks can be categorized based on when they are applied \cite{dziedzic2022difficulty}. Perturbation-based defenses, e.g., adding noise \cite{orekondy2019knockoff}, top-$k$ \cite{orekondy2019knockoff} and truncating outputs \cite{tramer2016stealing}, are applied before a stealing attack happens, aiming to limit the information leakage. 
On the other hand, watermarking defenses embed watermarks or unique identifiers into the target model and use them to detect whether it is stolen after a stealing attack happens \cite{jia2021entangled}.

\section{Methodology}

\subsection{Threat Model} \label{sec:threat model}
Given a target encoder $E_T$, the attacker aims to train a surrogate encoder $E_S$ at the lowest possible cost, which can perform competitively on downstream tasks with $E_T$. To achieve this goal, the attacker queries $E_T$ for embeddings of an unlabeled surrogated dataset $D_S=\{\boldsymbol{x}_1, \dots, \boldsymbol{x}_N\}$ consisting of $N$ images, and guide $E_S$ to mimic the output of $E_T$ for each sample in $D_S$. Specifically, we consider a black-box setting, where the attacker has direct access only to the outputs of $E_T$ while remaining unaware of its architecture and training configurations, including pre-training datasets, loss functions, data augmentations schemes, etc.

\subsection{Sample-Wise Prototypes}
Recalling the discussion in Section \ref{sec:intro}, we expect to refine the target encoder's representations to find a less biased optimization objective for each sample in $D_S$. Enlightened by prototype learning \cite{tan2022fedproto} and Extreme-Multi-Patch-SSL (EMP-SSL) \cite{tong2023emp}, which establish a prototype/benchmark for each class/sample, we propose a \textit{sample-wise prototype} generation method for this purpose. In detail, the process of prototype generation can be divided into three steps as follows: \ding{182} Cropping and augmenting each image $\boldsymbol{x}_i\in D_S$ into $n$ augmentation patches denoted as $\{\boldsymbol{x}'_{i,t,c}\}_{c=1}^{n} = \{\boldsymbol{x}'_{i,t,1}, \dots, \boldsymbol{x}'_{i,t,n}\}$; \ding{183} Querying $E_T$ with each augmentation patch in $\{\boldsymbol{x}'_{i,t,c}\}_{c=1}^{n}$, resulting in a set of embeddings denoted as $\{E_T(\boldsymbol{x}'_{i,t,c})\}_{c=1}^n = \{E_T(\boldsymbol{x}'_{i,t,1}), \dots, E_T(\boldsymbol{x}'_{i,t,n})\}$; \ding{184} Calculating the mean of $\{E_T(\boldsymbol{x}'_{i,t,c})\}_{c=1}^n$ as the prototype $p_{\boldsymbol{x}_i}$ for $\boldsymbol{x}_i$ as follows:
\begin{small}
\begin{equation} \label{eq:1}
    \resizebox{0.4\textwidth}{!}{$
    p_{\boldsymbol{x}_i} = \frac{1}{n}\sum_{c=1}^n E_T(\boldsymbol{x}'_{i,t,c}),\quad \boldsymbol{x}_i \in D_s.
    $}
\end{equation}
\end{small}

\begin{figure}[t]
    \centering
    \includegraphics[width=\textwidth]{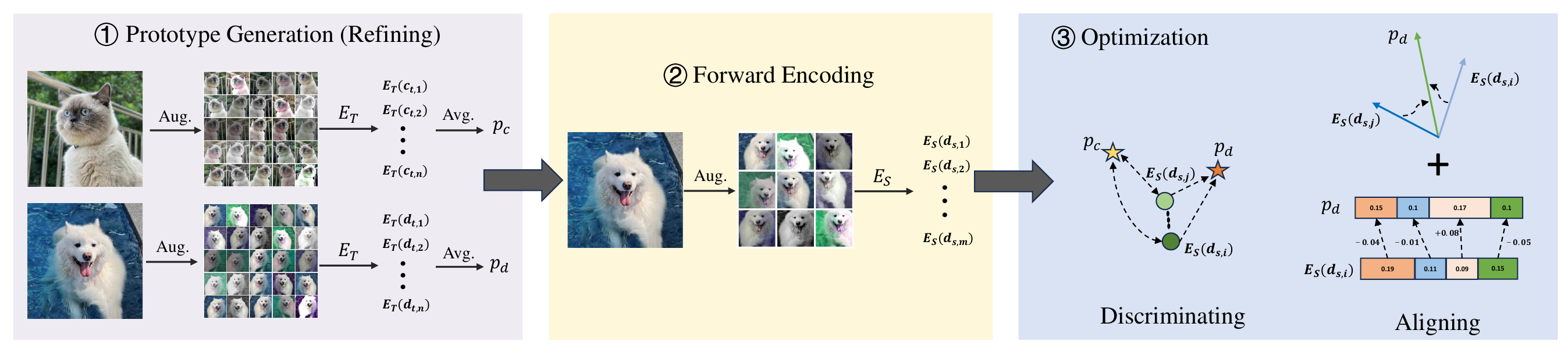}
    \caption{Pipeline of RDA. \textbf{Prototype generation}: augment one sample into $n$ patches and use them to query the target encoder ($E_T$). The mean of the $n$ patches's embeddings is defined as a prototype for this sample. \textbf{Forward encoding}: crop one image into $m$ patches and feed them to the surrogate encoder ($E_S$) for their embeddings. \textbf{Optimization:} align embeddings from the surrogate encoder to their matched prototypes in both angle and amplitude while pushing away those belonging to different samples.}
    \label{fig: pipeline}
\end{figure}

\noindent
Each generated prototype will be stored in a memory bank. These sample-wise prototypes can provide a stable optimization objective for each sample throughout the training process. Consequently, the attacker can accomplish the training in a query-free manner. Notably, setting $n$ to a value that is far less than the training epochs, e.g., 10 vs. 100, is sufficient to yield a favorable performance.
The prototype generation process is illustrated in step \ding{172} of Figure \ref{fig: pipeline}. Supplementary A.2 explains how we build the memory bank in practice.

\subsection{Multi-Relational Extraction Loss}
As step \ding{173} of Figure \ref{fig: pipeline} shows, during training $E_S$, we also perform cropping and augmentation on each image $\boldsymbol{x}_i \in D_S$ to create $m$ patches, forming an augmentation patch set denoted as $\{\boldsymbol{x}'_{i,s,q}\}_{q=1}^m = \{\boldsymbol{x}'_{i,s,1}, \dots, \boldsymbol{x}'_{i,s,m}\}$. The $m$ does not necessarily need to be set equal to $n$. Next, we feed each augmentation patch from $\{\boldsymbol{x}'_{i,s,q}\}_{q=1}^m$ to $E_S$, obtaining a set of embeddings denoted as $\{E_S(\boldsymbol{x}'_{i,s,q})\}_{q=1}^m = \{E_S(\boldsymbol{x}'_{i,s,1}),\dots, E_S(\boldsymbol{x}'_{i,s,m})\}$.  To optimize $E_S$, we develop a \textit{multi-relational extraction loss} composed of two parts, i.e., the \textit{discriminating loss} and the \textit{aligning loss}.

\noindent
\textbf{Discriminating Loss.}\quad The discriminating loss is responsible for training $E_S$ to distinguish between different samples, as depicted in step \ding{174} of Figure \ref{fig: pipeline}. For this goal, it pushes each embedding in $\{E_S(\boldsymbol{x}'_{i,s,q})\}_{q=1}^m$ away from mismatched prototypes $p_{\boldsymbol{x}_j}, i\neq j$ (denoted as negative pairs). In this regard, contrastive learning \cite{chen2020simple,khosla2020supervised,he2020momentum, chen2020improved} offers a solution, as it can pull embeddings of positive pairs close while pushing embeddings of negative pairs apart. Furthermore, since more negative pairs can help improve the performance of contrastive learning \cite{chen2020simple}, we refer to the loss designed by Sha \textit{et al.} \cite{sha2023can} to propose our discriminating loss $\mathcal{L}_D$. In particular, $\mathcal{L}_D$ considers both mismatched prototype-embedding pairs from different samples as well as embeddings from $E_S$ for different samples as negative pairs. Formally, $\mathcal{L}_D$ can be expressed as follows:
\begin{small}
\begin{equation} \label{eq:2}
    \resizebox{0.55\textwidth}{!}{$
    \mathcal{L}_{pos}(\boldsymbol{x}_i) = \frac{1}{m}\sum_{q=1}^m \exp{(sim(E_S(\boldsymbol{x}'_{i,s,q}), p_{\boldsymbol{x}_i})/\tau)},
    $}
\end{equation}
\end{small}
\begin{small}
\begin{equation} \label{eq:3}
\resizebox{0.55\textwidth}{!}{$
\begin{aligned}
\mathcal{L}_{neg}(\boldsymbol{x}_i)&= \frac{1}{m}\sum_{q=1}^m\sum_{k=1}^N\mathbbm{1}_{[i \neq k]}(\exp{(sim(E_S(\boldsymbol{x}'_{i,s,q}), p_{\boldsymbol{x}_k})/\tau)}\\ 
& \quad + \exp{(sim(E_S(\boldsymbol{x}'_{i,s,q}), E_S(\boldsymbol{x}'_{k,s,q}))/\tau})),
\end{aligned}
$}
\end{equation}
\end{small}
\begin{small}
\begin{equation}
  \resizebox{0.30\textwidth}{!}{$
  \mathcal{L}_{D} = -\frac{1}{N}\sum_{i=1}^N \log{\frac{\mathcal{L}_{pos}(\boldsymbol{x}_i)}{\mathcal{L}_{neg}(\boldsymbol{x}_i)}},
  $}
\end{equation}
\end{small}

\noindent
where $sim(u,v)$ represents the cosine similarity between $u$ and $v$, and $\tau$ is a temperature parameter.



\noindent
\textbf{Aligning Loss.}\quad While $\mathcal{L}_D$ can separate different samples effectively, it does not fully leverage the potential of prototypes. To further enhance the attack efficacy, we propose to align embeddings from $E_S$ more thoroughly with their matched prototypes from the high-performance $E_T$, i.e., in terms of both amplitude and angle, as step \ding{174} of Figure \ref{fig: pipeline} shows. To measure the amplitude and angle deviations between two embeddings, we employ the mean square error (MSE) and cosine similarity to quantify them respectively, which can be formulated as follows: 
\begin{small}
\begin{equation} \label{eq:5}
    \resizebox{0.45\textwidth}{!}{$
    \mathcal{L}'_{amp}(\boldsymbol{x}_i) = \frac{1}{m}\sum_{q=1}^{m} \Vert E_S(\boldsymbol{x}'_{i,s,q})-p_{\boldsymbol{x}_i}\Vert^2,
    $}
\end{equation}
\end{small}

\begin{small}
\begin{equation} \label{eq:6}
    \resizebox{0.5\textwidth}{!}{$
    \mathcal{L}'_{ang}(\boldsymbol{x}_i) = \frac{1}{m}\sum_{q=1}^{m} sim(E_S(\boldsymbol{x}'_{i,s,q}), p_{\boldsymbol{x}_i}).
    $}
\end{equation}
\end{small}

\noindent
Notice that we will normalize the value of $\mathcal{L}'_{ang}$ to $\left[0,1\right]$.
However, there is an issue concerning the equal penalty given by $\mathcal{L}'_{amp}$ and $\mathcal{L}'_{ang}$ to every deviation increase. For instance, when the MSE increases from 0.8 to 0.9 or from 0.3 to 0.4, both penalties given by $\mathcal{L}'_{amp}$ are 0.1, which is imprudent. On the contrary, we should have a more rigorous penalty for the increase in MSE from 0.3 to 0.4, as a smaller MSE is more desirable. Likewise, the same penalizing regime should be applied on $1/\mathcal{L}'_{ang}$ for the same reason.
To this end, we adopt a logarithmic function to redefine the ultimate formulations of $\mathcal{L}_{amp}$ and $\mathcal{L}_{ang}$ as follows:
\begin{equation} \label{eq:L_amp}
    \resizebox{0.3\textwidth}{!}{$
    \mathcal{L}_{amp}(\boldsymbol{x}_i) = \log \mathcal{L}'_{amp}(\boldsymbol{x}_i),
    $}
\end{equation}
\begin{equation} \label{eq:L_ang}
    \resizebox{0.5\textwidth}{!}{$
    \mathcal{L}_{ang}(\boldsymbol{x}_i) = \log (1/\mathcal{L}'_{ang} (\boldsymbol{x}_i)) = -\log \mathcal{L}'_{ang}(\boldsymbol{x}_i).
    $}
\end{equation}
We posit the amplitude and angle deviations hold equal importance, and formulate the aligning loss as follows:
\begin{small}
\begin{equation}
\resizebox{0.65\textwidth}{!}{$
\begin{aligned}
      \mathcal{L}_A &= \frac{1}{N}\sum_{i=1}^N(\mathcal{L}_{amp}(\boldsymbol{x}_i)+\mathcal{L}_{ang}(\boldsymbol{x}_i))
       =-\frac{1}{N}\sum_{i=1}^N \log \frac{\mathcal{L}'_{ang}(\boldsymbol{x}_i)}{\mathcal{L}'_{amp}(\boldsymbol{x}_i)}.
\end{aligned}
$}
\end{equation} 
\end{small}
Our experiments in Supplementary A.4 demonstrate the combination of $\mathcal{L}_{amp}$ and $\mathcal{L}_{ang}$ offers superior results than using each solely.

Lastly, the loss function for stealing pre-trained encoders is as follows:
\begin{equation} \label{eq:8}
    \resizebox{0.25\textwidth}{!}{$
    \mathcal{L} = \lambda_1 \cdot \mathcal{L}_{D} + \lambda_2 \cdot \mathcal{L}_{A},
    $}
\end{equation}
where $\lambda_1$ and $\lambda_2$ are preset coefficients to adjust the weight of each part in $\mathcal{L}$. 
To show the superiority of our design, especially on the loss function, we have conducted ablation studies in Section \ref{sec: ablation studies} and explored several alternative designs in Supplementary A.3. 

Detailed steps of RDA are summarized in Algorithm 1 of Supplementary A.2.

\section{Experiments}
\subsection{Experimental Setup}
\textbf{Target Encoder Settings.}\quad We use SimCLR \cite{chen2020simple} to pre-train two ResNet18 \cite{he2016deep} encoders on CIFAR10 \cite{krizhevsky2009learning} and STL10 \cite{coates2011analysis}, respectively, as two medium-scale target encoders. Furthermore, we consider two real-world large-scale ResNet50 encoders as the targets, i.e., the ImageNet encoder pre-trained by Google \cite{chen2020simple}, and the CLIP encoder pre-trained by OpenAI \cite{radford2021learning}. Besides the ResNet family, RDA also has been demonstrated effective upon other backbones, i.e., VGG19\_bn \cite{simonyan2014very}, DenseNet121 \cite{huang2017densely}, and MobileNetV2 \cite{howard2017mobilenets}, in Supplementary A.4.

\noindent
\textbf{Attack Settings.}\quad Regarding the surrogate dataset, it is derived from Tiny ImageNet \cite{le2015tiny}. Specifically, we randomly sample 2,500 images and resize them to $32\times32$ for stealing encoders pre-trained on CIFAR10 and STL10. When stealing the ImageNet encoder and CLIP, the number is 40,000 and 60,000, respectively, with each image resized to $224\times224$. For the surrogate encoder architecture, we adopt a ResNet18 across our experiments. During the attack, we set $n$ in Eq. \ref{eq:1} as 10, $m$ in Eq. \ref{eq:2}, \ref{eq:3}, \ref{eq:5}, and \ref{eq:6} as 5, $\tau$ in Eq. \ref{eq:2} and \ref{eq:3} as 0.07, and both $\lambda_1$ and $\lambda_2$ in Eq. \ref{eq:8} as 1 unless stated otherwise. For training, the batch size is set as 100, and we employ an Adam optimizer \cite{kingma2014adam} with a learning rate of 0.001.

\noindent
\textbf{Evaluation Settings.}\quad We train each surrogate encoder for 100 epochs and test its KNN accuracy \cite{wu2018unsupervised} on CIFAR10 after each epoch. The best-trained surrogate encoders, as well as target encoders, will be used to train downstream classifiers for the linear probing evaluation. Each downstream classifier will be trained for 100 epochs with an Adam optimizer and a learning rate of 0.0001.
Specifically, we totally consider seven downstream datasets, namely MNIST \cite{lecun-mnisthandwrittendigit-2010}, CIFAR10 \cite{krizhevsky2009learning}, STL10 \cite{coates2011analysis}, GTSRB \cite{stallkamp2012man}, CIFAR100 \cite{krizhevsky2010cifar}, SVHN \cite{netzer2011reading}, and F-MNIST \cite{xiao2017fashion}. Additional small-scale experiments on more complex datasets (e.g., Food 101 \cite{bossard2014food}) are included in our supplementary material. As done in \cite{liu2022stolenencoder}, we use Target Accuracy (TA) to evaluate target encoders, Steal Accuracy (SA) to evaluate surrogate encoders, and $\frac{\text{SA}}{\text{TA}}\times100\%$ to evaluate the efficacy of the stealing attack. 

\begin{table*}[t]
\centering
\begin{minipage}[c]{0.45\textwidth}
\centering
\captionof{table}{Results of RDA against two medium-scale encoders.} 
\label{tab: steal cifar10 and stl10}
\resizebox{0.9\textwidth}{!}{\begin{tabular}{ccccc}
\Xhline{1pt}
\multirow{2}{*}{\begin{tabular}[c]{@{}c@{}}Pre-training\\ Dataset\end{tabular}} &
  \multirow{2}{*}{\begin{tabular}[c]{@{}c@{}}Downstream\\ Dataset\end{tabular}} &
  \multirow{2}{*}{TA} &
  \multirow{2}{*}{SA} &
  \multirow{2}{*}{$\frac{\text{SA}}{\text{TA}}\times 100\%$} \\
                  &  &  &  &  \\ \Xhline{0.65pt}
\multirow{4}{*}{CIFAR10} & MNIST & 97.27 & 96.62 & 99.33  \\ 
                  & F-MNIST & 88.58 & 89.32 & 100.84 \\ 
                  & GTSRB & 61.76 & 62.75 & 101.60 \\ 
                  & SVHN & 73.78 & 73.74 & 99.95 \\ \Xhline{0.65pt}
\multirow{4}{*}{STL10} & MNIST & 96.84 & 96.31 & 99.45 \\ 
                  & F-MNIST & 90.08 & 87.91 & 97.59 \\ 
                  & GTSRB & 64.45 & 57.70 & 89.53 \\ 
                  & SVHN & 61.85 & 70.67  & 114.26  \\ \Xhline{1pt}
\end{tabular}}
\end{minipage}
\begin{minipage}[c]{0.45\textwidth}
\centering
\captionof{table}{Results of RDA against the ImageNet Encoder and CLIP.}
\label{tab: steal imagenet and clip}
\resizebox{0.85\textwidth}{!}{\begin{tabular}{ccccc}
\Xhline{1pt}
\multirow{2}{*}{\begin{tabular}[c]{@{}c@{}}Target\\ Encoder\end{tabular}} &
  \multirow{2}{*}{\begin{tabular}[c]{@{}c@{}}Downstream\\ Dataset\end{tabular}} &
  \multirow{2}{*}{TA} &
  \multirow{2}{*}{SA} &
  \multirow{2}{*}{$\frac{\text{SA}}{\text{TA}}\times 100\%$} \\
                  &  &  &  &  \\ \Xhline{0.65pt}
\multirow{4}{*}{{\begin{tabular}[c]{@{}c@{}}ImageNet\\Encoder\end{tabular}}} & MNIST & 97.38 & 95.89 &  98.47 \\ 
                  & F-MNIST & 91.67 & 90.83 & 99.08\\
                  & GTSRB & 63.14 &  59.62 & 94.43 \\ 
                  & SVHN & 73.20 & 69.21 & 94.55 \\ \Xhline{0.65pt}
\multirow{4}{*}{CLIP} & MNIST & 97.90 & 93.96 & 95.98 \\ 
                  & F-MNIST & 89.92 & 87.43 & 97.23 \\ 
                  & GTSRB & 68.61 & 55.84 & 81.39 \\ 
                  & SVHN & 69.53 & 57.58  & 82.81  \\ \Xhline{1pt}
\end{tabular}}
\end{minipage}
\end{table*}

\begin{table}[t]
\centering
\caption{\begin{small}
    Computation resources required by pre-training the two targeted real-world encoders from scratch and RDA to steal them.
\end{small}}
\label{tab: time of steal imagenet and clip}
\resizebox{0.45\textwidth}{!}{\begin{tabular}{cccccc}
\Xhline{1pt}
\multirow{2}{*}{\begin{tabular}[c]{@{}c@{}}Target\\ Encoder\end{tabular}} & \multicolumn{2}{c}{Pre-training} && \multicolumn{2}{c}{RDA}              \\ \cline{2-3} \cline{5-6} 
     & \multicolumn{1}{c}{Hardware} & Time (hrs) && \multicolumn{1}{c}{Hardware} & Time (hrs) \\ \Xhline{0.65pt}
\begin{tabular}[c]{@{}c@{}}ImageNet\\ Encoder\end{tabular}                & \multicolumn{1}{c}{TPU v3}   & 192 & & \multicolumn{1}{c}{\multirow{2}{*}{RTX A5000}} & 10.8 \\ 
CLIP & \multicolumn{1}{c}{V100 GPU} &  255,744 &        & \multicolumn{1}{c}{}         &   16.2     \\ \Xhline{1pt}
\end{tabular}}
\end{table}

\subsection{Effectiveness of RDA} \label{sec: effectiveness experiments}
\subsubsection{Stealing Medium-Scale Encoders.}
Table \ref{tab: steal cifar10 and stl10} shows the results achieved by RDA upon two medium-scale target encoders pre-trained on CIFAR10 and STL10, respectively. As the results show, using a surrogate dataset of 5\% the size of CIFAR10 (2,500 / 50,000) and 2.4\% the size of STL10 (2,500 / 105,000), RDA can steal nearly $100\%$ of the functionality of both target encoders, except for the GTSRB scenario when stealing the encoder pre-trained on STL10. Nonetheless, even in this exceptional case, the obtained $\frac{\text{SA}}{\text{TA}}\times100\%$ is approximately 90\%, which exemplifies the remarkable effectiveness of RDA. Furthermore, the surrogate encoder trained through RDA even outperforms the target encoder on multiple datasets. Our interpretation of this is that the target encoder fits the pre-training dataset more strongly compared to the RDA-trained surrogate encoder, thus exhibiting inferior performance on some out-of-distribution datasets.


\begin{figure*}[t]
	\centering
	\begin{minipage}{0.4\textwidth}
		\centering
		\includegraphics[width=1\textwidth]{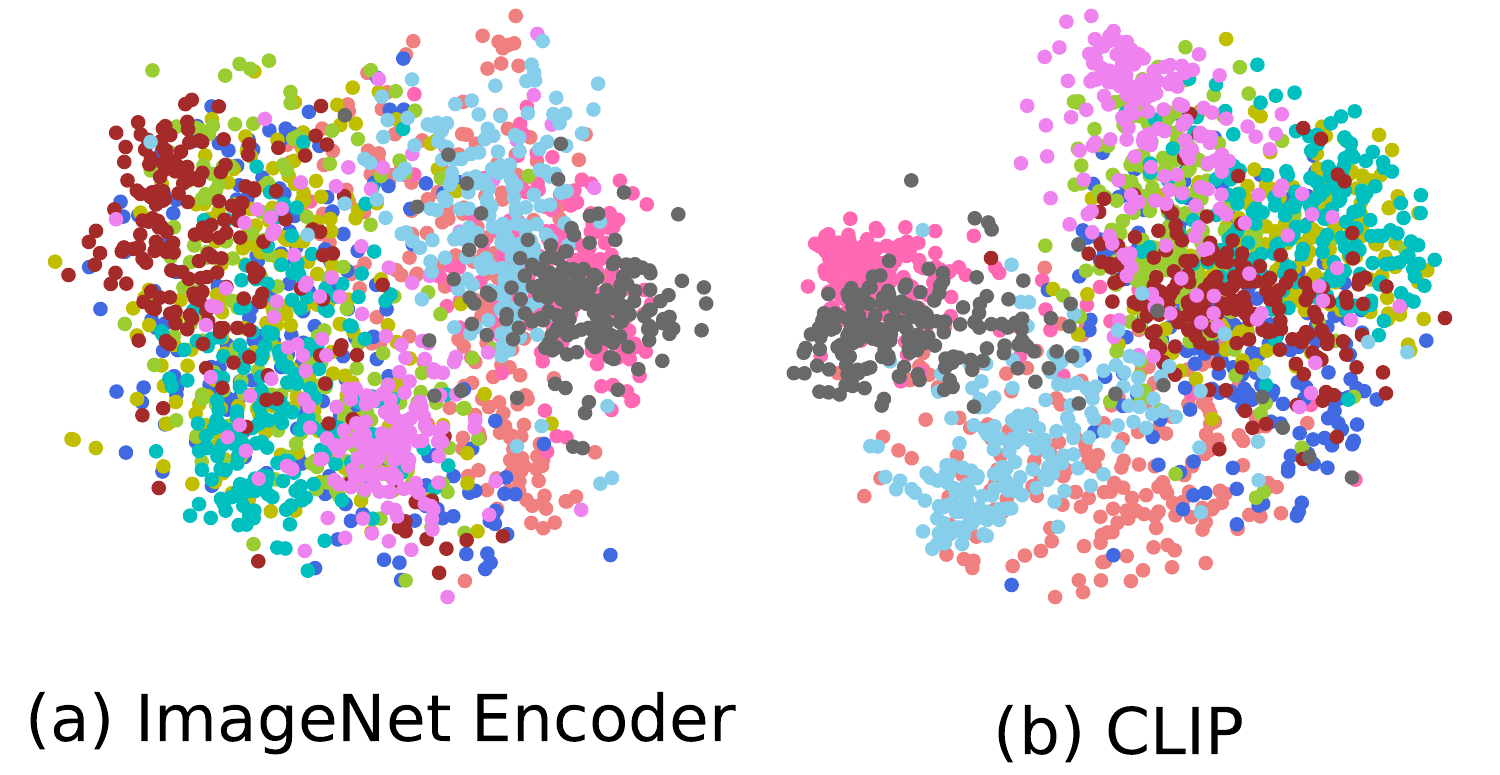}
		\caption{t-SNE of embeddings of 2,000 images sampled from CIFAR10 generated by the ImageNet encoder and CLIP.}
		\label{fig: tsne of imagenet and clip}
	\end{minipage}
	\begin{minipage}{0.58\textwidth}
		\centering
		\includegraphics[width=0.68\textwidth]{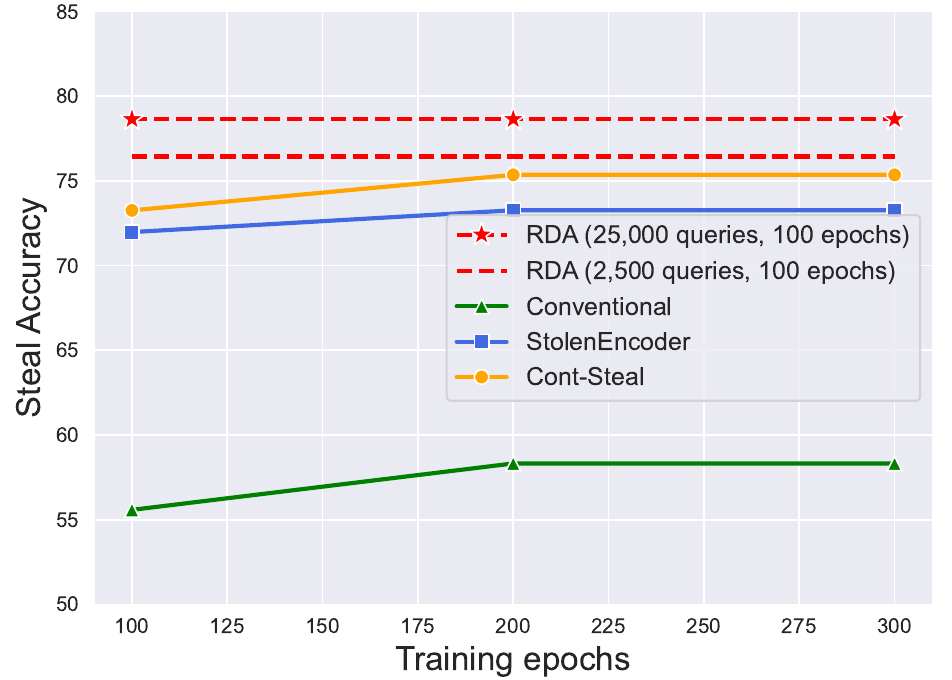}
		\caption{The training epochs of RDA are fixed
        to \textbf{100} while those of baselines are prolonged to \textbf{300}.}
		\label{fig: same training time}
	\end{minipage}
\end{figure*}


\subsubsection{Stealing Real-World Large-Scale Encoders.}
We further evaluate RDA on two real-world pre-trained encoders, i.e., the ImageNet encoder and CLIP. We aim to demonstrate the effectiveness of RDA upon such large-scale encoders trained with a massive amount of data. Specifically, the ImageNet encoder is pre-trained on the ImageNet dataset, which has 1.3 million images. The CLIP encoder is pre-trained on a web-scale dataset that consists of 400 million image-text pairs. During the attack, we set the patch number for training as 1 (i.e., $m$ in Eq. \ref{eq:2}, \ref{eq:3}, \ref{eq:5}, and \ref{eq:6}) to save time. Other parameters follow the default setting except the $\lambda_1$ and $\lambda_2$ when stealing CLIP, in which we set $\lambda_1$ and $\lambda_2$ to 1 and 20, respectively. To interpret the reason for this adjustment, we randomly sample 200 images from each class in the testing set of CIFAR10 and visualize the output embeddings from the ImageNet encoder and CLIP using t-SNE \cite{van2008visualizing}. Figure \ref{fig: tsne of imagenet and clip} shows that embeddings from the ImageNet encoder are more uniformly distributed compared to those from CLIP. This is because CLIP is a multimodal encoder, which is pre-trained by conducting contrastive learning between image-text pairs. Therefore, the text also shares part of the embedding space and is mutually exclusive with mismatched images. Although $\mathcal{L}_D$ can push embeddings of different samples apart, a significant weight of it will make all embeddings more uniformly distributed, i.e., not align with CLIP. Therefore, we set a smaller weight to $\mathcal{L}_D$ when stealing CLIP. This adjustment is practical since the attacker has access to the outputs of the target encoder, and thus can observe its embedding space to adjust its attack settings. Table \ref{tab: steal imagenet and clip} shows that using a surrogate dataset of only 3.07\% the size of ImageNet (40K / 1.3M) and 0.015\% the size of the training data of CLIP (60K / 400M), RDA can achieve comparable performances with the two encoders across various downstream tasks. The results also reveal that even under the architecture of the surrogate and target encoders being distinct (ResNet18 vs. ResNet50), RDA still exhibits high effectiveness. Moreover, Table \ref{tab: time of steal imagenet and clip} shows that RDA requires much fewer computation resources to steal the two encoders than pre-training them from scratch, owing to the small surrogate dataset and lightweight network architecture.

\begin{table*}[t]
\centering
\caption{Comparisons between RDA and baselines to steal the encoder pre-trained on CIFAR10 under the same surrogate dataset size. We report the \textbf{$\frac{\text{SA}}{\text{TA}}\times 100\%$} achieved by each method and its query cost. The \colorbox[RGB]{219,216,236}{Optimal} and \colorbox[RGB]{219,236, 212}{suboptimal} results are highlighted.}
\label{tab:4}
\begin{small}
\resizebox{0.7\textwidth}{!}{\begin{tabular}{ccccccccc}
\Xhline{1pt}
Method        & CIFAR10 & CIFAR100 & MNIST & GTSRB & SVHN & STL10 & F-MNIST & Queries\\ \Xhline{0.65pt}
Conventional \cite{sha2023can}  &  63.36  & 45.40 &96.16& 20.34  & 68.26  &60.25& 94.52  &\colorbox[RGB]{219,216,236}{2,500}\\ 
StolenEncoder \cite{liu2022stolenencoder} & 79.03 & 68.78 &98.76& 63.70 & 92.01& 76.29 & \colorbox[RGB]{219,236, 212}{95.91}&\colorbox[RGB]{219,216,236}{2,500} \\ 
Cont-Steal \cite{sha2023can}    & 80.43 &  73.94 &98.03&  \colorbox[RGB]{219,236, 212}{95.55} & 96.39   & 80.08 &  95.39 & 250,000 \\ \Xhline{0.65pt}
\multirow{2}{*}{RDA} & \colorbox[RGB]{219,236, 212}{83.84} & \colorbox[RGB]{219,236, 212}{77.49} &\colorbox[RGB]{219,236, 212}{98.97}& 94.27  & \colorbox[RGB]{219,236, 212}{96.79} & \colorbox[RGB]{219,236, 212}{82.25} & 94.69 & \colorbox[RGB]{219,216,236}{2,500} \\ 
 &  \colorbox[RGB]{219,216,236}{85.85} & \colorbox[RGB]{219,216,236}{83.08} & \colorbox[RGB]{219,216,236}{99.33} &\colorbox[RGB]{219,216,236}{101.60} & \colorbox[RGB]{219,216,236}{99.95}  & \colorbox[RGB]{219,216,236}{85.56}& \colorbox[RGB]{219,216,236}{100.84} & \colorbox[RGB]{219,236, 212}{25,000} \\ \Xhline{1pt}
\end{tabular}}
\end{small}
\end{table*}

\subsection{Comparision with Existing Methods}
We consider the \textit{conventional method} (CVPR 2023, proposed as the baseline by \cite{sha2023can}), \textit{StolenEncoder} (CCS 2022) \cite{liu2022stolenencoder}, and \textit{Cont-Steal} 
(CVPR 2023) \cite{sha2023can} as our baselines. A more detailed explanation of them is in Supplementary A.1. Specifically, we compare RDA with them to steal the encoder pre-trained on CIFAR10 under three different scenarios. For a fair comparison, the surrogate dataset remains identical across all baselines (except for some results presented in Table \ref{tab:5} due to different sizes of surrogate datasets).

\begin{small}
\begin{table*}[t]
\centering
\caption{Comparisons between RDA and StolenEncoder under the same query cost.}
\label{tab:5}
\resizebox{0.42\textwidth}{!}{\begin{tabular}{cccc}
\Xhline{1pt}
\begin{tabular}[c]{@{}c@{}}Queries\end{tabular} & Method       & setting                & SA \\ \hline
\multirow{5}{*}{2500}                                       & StolenEnoder & 2,500 images            &  71.98  \\ \cline{2-4} 
                                                            &\multirow{4}{*}{RDA}         & 500 images$\times$5 patches &  65.38  \\ 
                                                             & & 625 images$\times$4 patches &  67.07 \\ 
                                                             & & 1,250 images$\times$2 patches &  71.44 \\ 
                                                            &          & 2,500 images$\times$ 1 patch &  \colorbox[RGB]{219,216,236}{76.36}  \\ \Xhline{0.65pt}
\multirow{2}{*}{25,000}                                     & StolenEnoder & 25,000 images   & 76.23   \\ \cline{2-4} 
                                                            & RDA        & 2,500 images$\times$10 patches  &   \colorbox[RGB]{219,216,236}{78.50} \\ \Xhline{1pt}
\end{tabular}}
\end{table*}
\end{small}


\subsubsection{Under the Same Surrogate Dataset Size.}
In this subsection, we fix the surrogate dataset size to 2,500 and make it identical across all methods. As Table \ref{tab:4} shows, RDA outperforms baselines by a significant margin across all seven downstream datasets with a moderate query cost. We also evaluate RDA under a setting that has the same query cost as the conventional method and StolenEncoder (i.e., 2,500 queries), for which we query with each sample once in its origin version. The results show that RDA still surpasses baselines over five out of seven datasets with the least query cost. In addition, the comparison between the results achieved by RDA with 25,000 and 2,500 queries reveals that more patches for generating prototypes will enhance the attack. 
The embedding space of the surrogate encoder trained by each method is visualized in Figure 11 of Supplementary A.4, which shows that our RDA can train surrogate encoders to discriminate different samples better. Besides, We have summarized the mean value achieved by each method over the seven downstream datasets in Figure \ref{fig: results summary} to show the superiority of RDA.

\begin{figure*}[t]
    \centering
    \includegraphics[width=\textwidth]{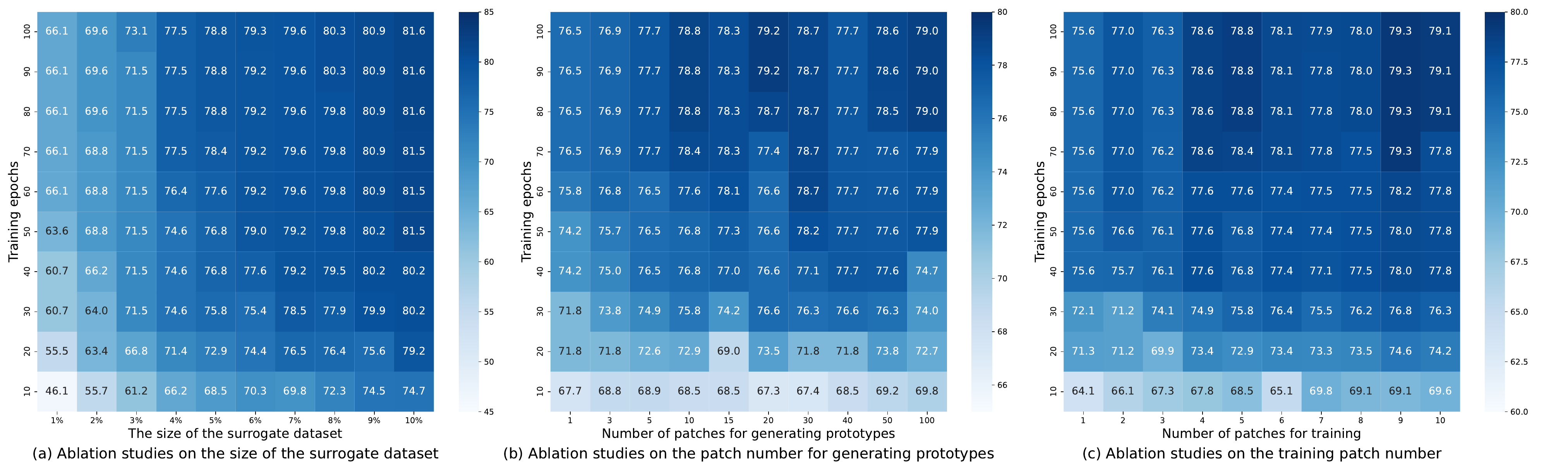}
    \caption{Heat map of the achieved SAs (better zoom in). This figure shows the performance of 100 combinations of different training epochs and (a) the surrogate dataset size, (b) the patch number for generating prototypes, and (c) the patch number for training.}
    \label{fig: heat map}
\end{figure*}


\subsubsection{Under the Same Query Cost.}
To demonstrate the superiority of RDA against baselines under an identical query cost, we consider two fixed values, i.e., 2,500 and 25,000. We take StolenEncoder as the rival since it is the best baseline with the least query cost. We note that Cont-Steal is beyond our consideration because the data volume for it would be too small under such a setting, which is only 25 images for 2,500 queries and 250 images for 25,000 queries. $N$ images $\times$ $n$ patches in the table indicates that we use $N$ images as the surrogate dataset and crop each image into $n$ patches to generate prototypes. Therefore, the query cost of RDA is $N \times n$. As shown in Table \ref{tab:5}, under the same query cost of 2,500, RDA can achieve comparable results with StolenEncoder with half the size of its surrogate dataset. With the same 2,500 images, RDA outperforms StolenEncoder by a significant margin, showcasing its superiority. Moreover, under a query cost of 25,000, RDA outperforms StolenEncoder by 2.14\% with only 10\% the size of its surrogate dataset. Besides, comparisons between different configurations of RDA under queries of 2,500 indicate that the training data volume has a dominant impact on the performance.

\subsubsection{Under the Same Time Cost.}
Since each sample is also cropped and augmented into multiple patches for training (i.e., needing to forward encode each sample multiple times), RDA takes the longest time for training. Table 10 in Supplementary A.4 contains the detailed time cost of each method. To make a fair comparison, we prolong the training for another 200 epochs for baseline methods, i.e., 300 epochs in total, which is sufficient to lead their performances to saturate. The trained surrogate encoders are then evaluated on CIFAR10 as the downstream dataset. Figure \ref{fig: same training time} reveals that even though baseline methods take longer time to train, RDA still outperforms them with the least query cost, and the performance gap can be further widened by a moderate query cost.

\begin{figure*}[t]
    \centering
    \includegraphics[width=0.85\textwidth]{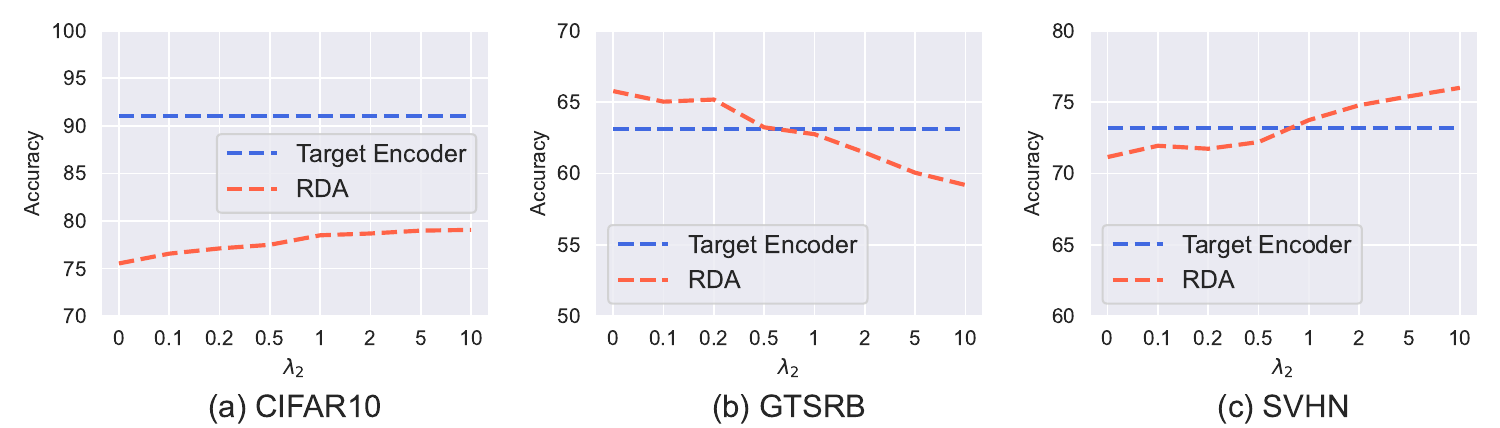}
    \caption{Ablation studies on the weight of each part in $\mathcal{L}$, with $\lambda_1 = 1$ and $\lambda_2$ varying.}
    \label{fig: ablations on the loss weight}
\end{figure*}

\subsection{Ablation Studies} \label{sec: ablation studies}
In this subsection, we conduct ablation studies to investigate the impact of the surrogate dataset size, the patch number for generating prototypes (i.e., $n$ in Eq. \ref{eq:1}), the patch number for training (i.e., $m$ in Eq. \ref{eq:2}, \ref{eq:3}, \ref{eq:5}, and \ref{eq:6}), and loss functions. All the experiments are conducted to steal the encoder pre-trained on CIFAR10. For evaluations, the trained surrogate encoders are assessed on CIFAR10 in the first three ablation studies. While examining the influence of loss functions, we consider seven distinct downstream datasets, as shown in Table \ref{tab:4}, to derive a more comprehensive conclusion. Other configurations remain at their default settings, and we present the resulting SAs.

\subsubsection{Surrogate Dataset Size.} 
Recall our default setting where we randomly sample 2,500 images from Tiny ImageNet to construct the surrogate dataset, which is 5\% the size of CIFAR-10. We vary the ratio from 1\% to 10\% to investigate its impact. Figure \ref{fig: heat map} (a) shows that a larger surrogate dataset will accelerate the training and improve the attack performance.

\subsubsection{Patch Number for Generating Prototypes.}
We vary the patch number for generating prototypes from 1 to 100 to investigate its impact. Figure \ref{fig: heat map} (b) shows that there is a tendency for more patches for generating prototypes will improve the attack performance and accelerate the convergence. However, when the patch number reaches 10, more patches do not further improve the attack performance due to the limited surrogate dataset size.

\begin{table}[t]
\centering
\caption{Ablation studies on loss functions. The presented results are SAs.}
\label{tab: loss}
\begin{small}
\resizebox{0.7\textwidth}{!}{\begin{tabular}{ccccccccc}
\Xhline{1pt}
Loss Type       &  CIFAR10 & STL10 & CIFAR100 & MNIST & F-MNIST & GTSRB & SVHN & AVG \\ \Xhline{0.65pt}
MSE               & 65.66 & 53.83 &  31.00 & 94.63 & 86.12     &  32.79     &   62.97 & 61.00   \\ 
Cosine Similarity &  80.23 & 66.56 & 43.81 & \colorbox[RGB]{219,216,236}{97.00} & 88.76 &   55.53    &  75.94 & 72.54  \\ 
KL Divergence     &77.35 & 65.30 &   40.98  & 96.23  &    87.82   &   57.04    &   69.38 & 70.58 \\ 
InfoNCE           & 73.99 & 61.68 &   39.24 &  95.76  &  87.86     &   58.82    &   68.88  & 69.43\\ \Xhline{0.65pt}
$\mathcal{L}_D$     & 75.54 & 63.2 &     43.16 & 96.52  &   89.27    &   \colorbox[RGB]{219,216,236}{65.77}    &   71.15  & 72.23\\ 
$\mathcal{L}_A$     & \colorbox[RGB]{219,216,236}{79.70} & 65.59 &    43.79  & 96.89  &   89.24    &   55.96    &  \colorbox[RGB]{219,216,236}{76.70}  &72.55  \\ 
$\mathcal{L}_A+\mathcal{L}_D$ & 79.39 & \colorbox[RGB]{219,216,236}{66.76} &  \colorbox[RGB]{219,216,236}{44.27} & 96.74 &  \colorbox[RGB]{219,216,236}{89.32}   &   62.75    &   73.74 & \colorbox[RGB]{219,216,236}{73.28} \\ \Xhline{1pt}
\end{tabular}} 
\end{small}
\end{table}

\subsubsection{Patch Number for Training.}
We vary the patch number for training from 1 to 10 to investigate its impact. Figure \ref{fig: heat map} (c) reveals a tendency for more patches for training will accelerate the training and achieve better results. However, more patches also indicate longer training time and more computation resources.

\subsubsection{Loss Functions.}
To demonstrate the superiority of the design of our loss function and the necessity of each part, we conduct ablation studies as shown in Table \ref{tab: loss}. Table \ref{tab: loss} shows that $\mathcal{L}_D$ largely surpass other loss functions on GTSRB while $\mathcal{L}_A$ largely surpass other loss functions on SVHN.
We hypothesize that $\mathcal{L}_A$ attains optimal outcomes on SVHN due to the pronounced resemblance between SVHN and the surrogate dataset. It appears that $\mathcal{L}_A$ inclines towards inducing the surrogate encoder to overfit the surrogate data, thereby elucidating its suboptimal performance on GTSRB, a dataset characterized by less similarity with the surrogate dataset. On the contrary, $\mathcal{L}_D$ makes the surrogate encoder fit less to the surrogate dataset and has the best result on GTSRB. To further investigate the effect of each part in our loss, we have conducted an ablation experiment about the weight of each part in our loss. 
As Figure \ref{fig: ablations on the loss weight} shows, the surrogate encoder will perform better on SVHN while worse on GTSRB when the weight of $\mathcal{L}_{A}$ increases. We refer to Supplementary A.4 for a more detailed analysis. Combining $\mathcal{L}_D$ and $\mathcal{L}_A$, our loss is the most robust one, which achieves the best average result over the seven tested datasets. For example, although cosine similarity and $\mathcal{L}_A$ performs well on SVHN (about 76\%), their performance on GTSRB is terrible (about 56\%). Our loss design can improve their performance on GTSRB by about 7\% in the cost of less than 3\% decrease on SVHN.

\subsection{Robustness to Defenses}
In this section, we evaluate the robustness of RDA against three perturbation-based defenses that aim to limit information leakage and one watermarking defense that aims to detect whether a suspected encoder is stolen. The pre-training and downstream datasets both are set to CIFAR10.

\subsubsection{Perturbation-Based Defense.}
We evaluate three common practices of this type of defense, i.e., adding noise \cite{orekondy2019knockoff}, top-$k$ \cite{orekondy2019knockoff} and rounding \cite{tramer2016stealing}.

$\bullet$ \textbf{Adding noise:} Adding noise means that the defender will introduce noise to the original outputs of the target encoder. Following \cite{dziedzic2022difficulty}, we set the mean of the noise to 0 and vary the standard deviations to control the noise level. 

$\bullet$ \textbf{Top-$k$:} Top-$k$ means that the defender will only output the first $k$ largest number of each embedding from the target encoder and set the rest as 0. We vary the value of $k$ to simulate different perturbation levels.

$\bullet$ \textbf{Rounding:} Rounding truncates each value in the embedding to a specific precision, which we vary to simulate different perturbation levels. 

\begin{figure*}[t]
    \centering
    \includegraphics[width=\textwidth]{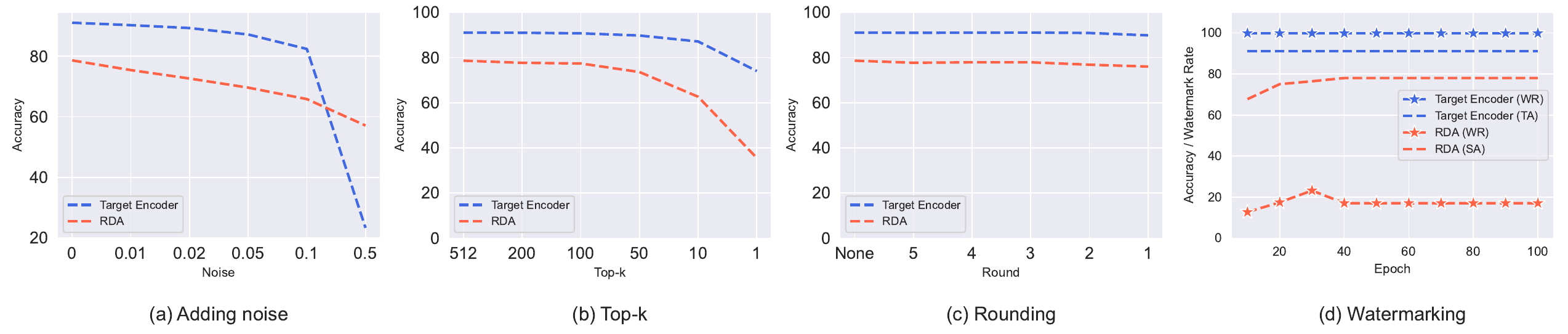}
    \caption{The performance of different defense methods on CIFAR10.}
    \label{fig: defense results}
\end{figure*}

The results of the three perturbation-based defense methods are summarized in (a), (b), and (c) of Figure \ref{fig: defense results}, respectively. We can observe in Figure \ref{fig: defense results} (a) that while adding noise can mitigate the model stealing attack, it decreases the utility of the target encoder more significantly. 
Moreover, Figure \ref{fig: defense results} (b) shows that though top-$k$ can decrease the efficacy of RDA as the defender lowers the value of $k$, it also severely deteriorates the target encoder's performance. On the other hand, Figure \ref{fig: defense results} (c) reveals that rounding has a minimal effect on the performance of both the target encoder and attack. Our experimental results demonstrate the robustness of RDA against perturbation-based defenses.

\subsubsection{Watermarking Defense.}  
    As shown by Adi \textit{et al.} \cite{adi2018turning}, backdoors can be used as watermarks to claim the ownership of a model. In this sense, we follow \cite{sha2023can} and leverage BadEncoder \cite{jia2022badencoder} to embed a backdoor into the target encoder as the watermark. Ideally, if the watermarked encoder is stolen, the surrogate encoder trained via stealing should also be triggered by the defender-specified trigger to exhibit certain behaviors, and thus the ownership can be claimed. Fortunately, Figure \ref{fig: defense results} (d) shows that RDA can steal the functionality of the target encoder while unlearning the watermark. More specifically, the watermark rate (WR) of the target encoder is 99.87\% but only 17.04\% for the surrogate encoder when the training converges. This observation indicates that the watermark embedded in the target encoder cannot be preserved by RDA. 


\section{Conclusion}
In this paper, we have proposed a novel model stealing method against SSL named RDA, which stands for \textit{refine}, \textit{discriminate}, and \textit{align}. Compared with previous methods \cite{liu2022stolenencoder, sha2023can}, RDA establishes a less biased optimization objective for each training sample and strives to extract more abundant functionality of the target encoder. This empowers RDA to exhibit tremendous effectiveness, efficiency, and robustness across diverse datasets and settings.

\noindent
\textbf{Ethical Concerns and Possible Defenses.} We underscore that the misuse of model stealing techniques can jeopardize the privacy and economic rights of embedding service providers. We hope that our work will inspire the development of more sophisticated defense mechanisms to thwart model stealing attacks. A possible defense involves rejection schemes for suspicious queries that appear highly similar, which we have more detailed discussions about in the Supplementary. 

\noindent
\textbf{Limitation and Future Work.} The query cost of RDA is not yet optimal and can be further reduced by techniques like clustering to establish an identical prototype for multiple similar images. Moreover, more unique (e.g., more efficient) designs of RDA for transformer-based huger models remain largely unexplored.

\subsubsection{Acknowledgements.}
This work is supported by the National Key R\&D Program of China under Grant 2022YFB3103500, the National Natural Science Foundation of China under Grant U20A20176 and 62072062, the National Natural Science Foundation of Chongqing under Grant cstc2022ycjh-bgzxm0031, and the Chongqing Research Program of Basic Research and Frontier Technology (Chongqing Talent) under Grant cstc2024ycjh-bgzxm0048.

%
%
\bibliographystyle{splncs04}
\bibliography{main}

\clearpage
\setcounter{page}{1}

\appendix{\section{Supplementary Material}}
\subsection{Elaborations on Existing Methods} \label{sup: representative methods}
In this section, we will elaborate on the three baselines considered in this paper, namely, the \textit{conventional method} \cite{sha2023can}, \textit{StolenEnoder} \cite{liu2022stolenencoder}, and \textit{Cont-Steal} \cite{sha2023can}.

$\bullet$ \textbf{Conventional method:} Given a surrogate dataset $D_S=\{\boldsymbol{x}_1,\boldsymbol{x}_2,\dots,\boldsymbol{x}_N\}$, the conventional method first queries the target encoder $E_T$ once with each sample $\boldsymbol{x}_i \in D_S$. The output embedding $E_T(\boldsymbol{x}_i), i=1,2,\dots,N,$ will be stored in a memory bank to serve as a ``ground truth'' for the sample across the whole training. Next, to train the surrogate encoder $E_S$, it feeds each sample $\boldsymbol{x}_i \in D_S$ to $E_S$ and pulls the output embedding $E_S(\boldsymbol{x}_i)$ close to its corresponding optimization objective, i.e., $E_T(\boldsymbol{x}_i)$. 

$\bullet$ \textbf{StolenEncoder:} Similar to the conventional method, StolenEncoder first queries the target encoder $E_T$ once with each sample $\boldsymbol{x}_i \in D_S$ and stores the output embedding $E_T(\boldsymbol{x}_i), i=1,2,\dots,N,$ in a memory bank. Next, it not only feeds each sample $\boldsymbol{x}_i \in D_S$ but also an augmentation $\boldsymbol{x}'_i$ of it to the surrogate encoder $E_S$. The output embeddings $E_S(\boldsymbol{x}_i)$ and $E_S(\boldsymbol{x}'_i)$ will be pulled close to $E_T(\boldsymbol{x}_i)$ simultaneously for optimizing the surrogate encoder $E_S$. 

$\bullet$ \textbf{Cont-Steal:} In particular, Cont-Steal adopts an end-to-end training strategy to train the surrogate encoder $E_S$. In each epoch, Cont-Steal first augments each image $\boldsymbol{x}_i \in D_S$ into two perspectives, denoted as $\boldsymbol{x}'_{i,t}$ and $\boldsymbol{x}'_{i,s}$, respectively. Next, $\boldsymbol{x}'_{i,t}$ is used to query the target encoder $E_T$, while $\boldsymbol{x}'_{i,s}$ is fed to the surrogate encoder $E_S$. The surrogate encoder then is optimized via contrastive learning that pulls $E_T(\boldsymbol{x}'_{i,t})$ and $E_S(\boldsymbol{x}'_{i,s})$ close while pushing $E_S(\boldsymbol{x}'_{k,s}), k \neq i,$ and $E_T(\boldsymbol{x}'_{i,t})$ apart. Moreover, to enhance contrastive learning, Cont-Steal also includes $E_S(\boldsymbol{x}'_{i,s})$ and $E_S(\boldsymbol{x}'_{k,s}), i \neq k,$ as negative pairs to train $E_S$. 

\begin{table}[h]
\centering
\caption{Query cost analysis of three baseline methods.}
\label{tab: query cost analysis}
\resizebox{0.6\textwidth}{!}{\begin{tabular}{cccc}
\Xhline{1pt}
Method & \begin{tabular}[c]{@{}c@{}}Surrogate Dataset\\ Size\end{tabular} & Training Epoch & Query Cost \\ \Xhline{0.65pt}
Conventioanl \cite{sha2023can} & $N$ & $L$ & $N$ \\ 
StolenEncoder \cite{liu2022stolenencoder} & $N$ & $L$ & $N$ \\ 
Cont-Steal \cite{sha2023can} & $N$ & $L$ & $N \times L$ \\ \Xhline{1pt}
\end{tabular}}
\end{table}

\noindent
\textbf{Query Cost Analysis.} Assuming the attacker has a surrogate dataset $D_S=\{\boldsymbol{x}_1,\boldsymbol{x}_2,\dots,\boldsymbol{x}_N\}$ consisting of $N$ images. To train a surrogate encoder, the attacker will use $D_S$ to query the target encoder and optimize the surrogate encoder to mimic its outputs. Regarding the conventional method and StolenEncoder, each image $\boldsymbol{x}_i \in D_s$ will be used to query only once. Therefore, the query cost of them is $N$, and nothing about the number of training epochs. As for Cont-Steal, each image in $D_S$ will be augmented into two perspectives and one is used to query the target encoder in each epoch. Therefore, assuming the training will last for $L$ epochs, the query cost of Cont-Steal is $N \times L$. Although Cont-Steal achieves superior results, its query cost is formidable since the training typically requires over 100 epochs to converge. We have the query cost of each method summarized in Table \ref{tab: query cost analysis}.

\begin{algorithm}[t]
\small
\caption{Detailed steps of RDA}
\label{alg:RDA}
\SetKwData{Left}{left}\SetKwData{This}{this}\SetKwData{Up}{up} \SetKwFunction{Union}{Union}\SetKwFunction{FindCompress}{FindCompress}
\SetKwInOut{Input}{input}\SetKwInOut{Output}{output}
\Input{Surrogate dataset $D_S$, target encoder $E_T$, surrogate encoder $E_S$, patch number $n$ for generating prototypes and $m$ for training.} 
\Output{A high-performance surrogate encoder $E_S$}

Load $E_T$ and initialize $E_S$.\\

\textbf{Prototype generation:}\\

\For{each image $\boldsymbol{x}_i \in D_S$}{
augment $\boldsymbol{x}_i$ into multiple patches $\{\boldsymbol{x}'_{i,t,c}\}_{c=1}^{n}=\{\boldsymbol{x}'_{i,t,1},\dots,\boldsymbol{x}'_{i,t,n}\}$;\\
query $E_T$ with $\{\boldsymbol{x}'_{i,t,c}\}_{c=1}^{n}$ and obtain $\{E_T(\boldsymbol{x}'_{i,t,c})\}_{c=1}^{n}=\{E_T(\boldsymbol{x}'_{i,t,1}),\dots,E_T(\boldsymbol{x}'_{i,t,n})\}$;\\
calculate the sample-wise prototype for $\boldsymbol{x}_i$ as follows: $p_{\boldsymbol{x}_i}=\frac{1}{n}\sum_{c=1}^n E_T(\boldsymbol{\boldsymbol{x}}'_{i,t,c})$;
}

\textbf{Training:}\\
\For{each epoch}{
\For{each image $\boldsymbol{x}_i \in D_S$}{
augment $\boldsymbol{x}_i$ into multiple patches $\{\boldsymbol{x}'_{i,s,q}\}_{q=1}^{m}=\{\boldsymbol{x}'_{i,s,1},\dots,\boldsymbol{x}'_{i,s,m}\}$;\\
feed $\{\boldsymbol{x}'_{i,s,q}\}$ into $E_S$ and obtain $\{E_S(\boldsymbol{x}'_{i,s,q})\}_{q=1}^{m}=\{E_S(\boldsymbol{x}'_{i,s,1}),\dots,E_S(\boldsymbol{x}'_{i,s,m})\}$;\\

\For{each $E_S(\boldsymbol{x}'_{i,s,1}) \in \{E_S(\boldsymbol{x}'_{i,s,q})\}_{q=1}^{m}$}{
optimize $E_S$ with $\mathcal{L}=\lambda_1 \cdot \mathcal{L}_D + \lambda_2 \cdot \mathcal{L}_A$ based on Eq. \ref{eq:2}-\ref{eq:8};\\  
}
}
}
\textbf{return:} the trained surrogate encoder $E_S$
\end{algorithm}

\subsection{Details about Our Method} \label{sup: details about our method}

\textbf{Practical Formulation of the Memory Bank.}\quad In practice, we assign a unique label to each image in the attacker's surrogate dataset as its key in the memory bank. Specifically, we transform $D_s=\{\boldsymbol{x}_1,\boldsymbol{x}_2,\dots,\boldsymbol{x}_N\}$ into a set of image-key pairs, i.e., $D_s=\{(\boldsymbol{x}_1,1),(\boldsymbol{x}_2,2),\dots,(\boldsymbol{x}_n,n)\}$. Then we generate a prototype for each image and get a set of key-prototype pairs that can be expressed as $P=\{1:p_{\boldsymbol{x}_1}, 2:p_{\boldsymbol{x}_2}, \dots, n:p_{\boldsymbol{x}_n}\}$. The prototype set $P$ is stored in a memory bank and we do not need to query the target model during training anymore.

\noindent
\textbf{Detailed Steps.}\quad Detailed steps of RDA are summarized in Algorithm \ref{alg:RDA}.

\begin{table}[t]
\centering
\caption{Results of different loss designs. The results are SAs on different downstream datasets, with highlighting the \colorbox[RGB]{219,216,236}{optimal}.}
\label{tab: loss designs in sup}
\resizebox{0.6\textwidth}{!}{\begin{tabular}{ccccc}
\Xhline{1pt}
Loss Type          & CIFAR100 & F-MNIST & GTSRB & SVHN \\ \Xhline{0.65pt}
$\mathcal{L}$ (current used) &    \colorbox[RGB]{219,216,236}{44.27}     &  \colorbox[RGB]{219,216,236}{89.32}   &   \colorbox[RGB]{219,216,236}{62.75}    &   \colorbox[RGB]{219,216,236}{73.74}  \\  \Xhline{0.65pt}
$\mathcal{L}_{v2}$         &  41.99  &  85.45  &  61.92  &  72.88 \\ 
$\mathcal{L}_{v3}$         &  42.23  &  88.63  &  62.46  &  71.86   \\ 
\Xhline{1pt}
\end{tabular}}
\end{table}

\begin{table}[t]
\centering
\caption{Ablation studies on the weights of $\mathcal{L}_{amp}$ and $\mathcal{L}_{ang}$. The results are SAs on different downstream datasets, with highlighting the \colorbox[RGB]{219,216,236}{optimal}.}
\label{tab: loss weight of amp and ang}
\resizebox{0.6\textwidth}{!}{\begin{tabular}{ccccc}
\Xhline{1pt}
$\mathcal{L}_{amp}:\mathcal{L}_{ang}$  & CIFAR100 & F-MNIST & GTSRB & SVHN \\ \Xhline{0.65pt}
$5:1$ &  42.90  &  86.52 & 61.32 &  70.57 \\  
$2:1$         &  43.61  & 86.80 & 62.55  & 72.90  \\ 
$1:1$         & \colorbox[RGB]{219,216,236}{44.27}  & \colorbox[RGB]{219,216,236}{89.32} & \colorbox[RGB]{219,216,236}{62.75}  &  \colorbox[RGB]{219,216,236}{73.74} \\ 
$1:2$         & 42.70  & 88.64 & 60.59  &  72.05 \\ 
$1:5$         & 43.24  & 89.16 & 62.04  & 73.11  \\ 
\Xhline{1pt}
\end{tabular}}
\end{table}
 
\subsection{Loss Designs} \label{sup: alternative loss designs}
\subsubsection{Alternative Loss Designs.}
In this section, we consider two alternative loss designs about $\mathcal{L}_A$. For simplicity, we denote them as $\mathcal{L}_{A,v2}$ and $\mathcal{L}_{A,v3}$, respectively.

$\bullet$ $\mathcal{L}_{A,v2}$: To investigate the benefits of penalizing different deviations with different levels, we define $\mathcal{L}_{A,v2}$ as a naive combination of $\mathcal{L}'_{amp}$ and $\mathcal{L}'_{ang}$ as follows:
\begin{equation}
    \mathcal{L}_{A,v2} = \frac{1}{N} \sum_{i=1}^N (\mathcal{L}'_{amp}(\boldsymbol{x}_i) - \mathcal{L}'_{ang}(\boldsymbol{x}_i)).
\end{equation}

$\bullet$ $\mathcal{L}_{A,v3}$: To investigate the benefits of our current penalizing regime employed by $\mathcal{L}_A$, we reverse it here by an ``$\exp$'', i.e., penalizing the MSE increase from 0.8 to 0.9 more harshly than that from 0.3 to 0.4, with the same rule employed on the reciprocal of the cosine similarity. Therefore, we formulate the $\mathcal{L}_{A,v3}$ as follows:
\begin{equation}
    \mathcal{L}_{A,v3} = \frac{1}{N} \sum_{i=1}^N (\exp{(\mathcal{L}'_{amp}(\boldsymbol{x}_i))} + \exp{(1/\mathcal{L}'_{ang}(\boldsymbol{x}_i))}).
\end{equation}
Finally, we define:
\begin{equation}
    \mathcal{L}_{v2} = \mathcal{L}_{D} + \mathcal{L}_{A,v2},
\end{equation}
\begin{equation}
    \mathcal{L}_{v3} = \mathcal{L}_{D} + \mathcal{L}_{A,v3}.
\end{equation}

We use the encoder pre-trained on CIFAR10 as the target encoder and evaluate the trained surrogate encoders on four different downstream datasets, i.e., CIFAR100, F-MNIST, GTSRB, and SVHN. Each experiment is run three times and we report the mean value of the achieved SAs by each loss design in Table \ref{tab: loss designs in sup}. 

As the results show, both $\mathcal{L}_{v2}$ and $\mathcal{L}_{v3}$ underperform $\mathcal{L}$. On the one hand, $\mathcal{L}$ outperforms $\mathcal{L}_{v2}$ indicating that incorporating our current penalizing regime is beneficial. On the other hand, $\mathcal{L}$ outperforms $\mathcal{L}_{v3}$ indicating that penalizing a deviation from a more favorable value to a worse one more harshly is more favorable upon the opposite penalizing regime.  

\subsubsection{The Weights of Amplitude and Angle Deviations.}
Recall our default setting that we assume the amplitude and angle deviations are of equal importance and thus formulate $\mathcal{L}_A(\boldsymbol{x}_i) = \mathcal{L}_{amp}(\boldsymbol{x}_i) + \mathcal{L}_{ang}(\boldsymbol{x}_i)$. Here we ablate on five different weight pairs as shown in Table \ref{tab: loss weight of amp and ang} to investigate the importance level of each deviation type. From the results, we can see that setting the weights of $\mathcal{L}_{amp}$ and $\mathcal{L}_{ang}$ to $1:1$ generally achieves the best result.

\begin{figure}[t]
    \centering
    \includegraphics[width=0.5\textwidth]{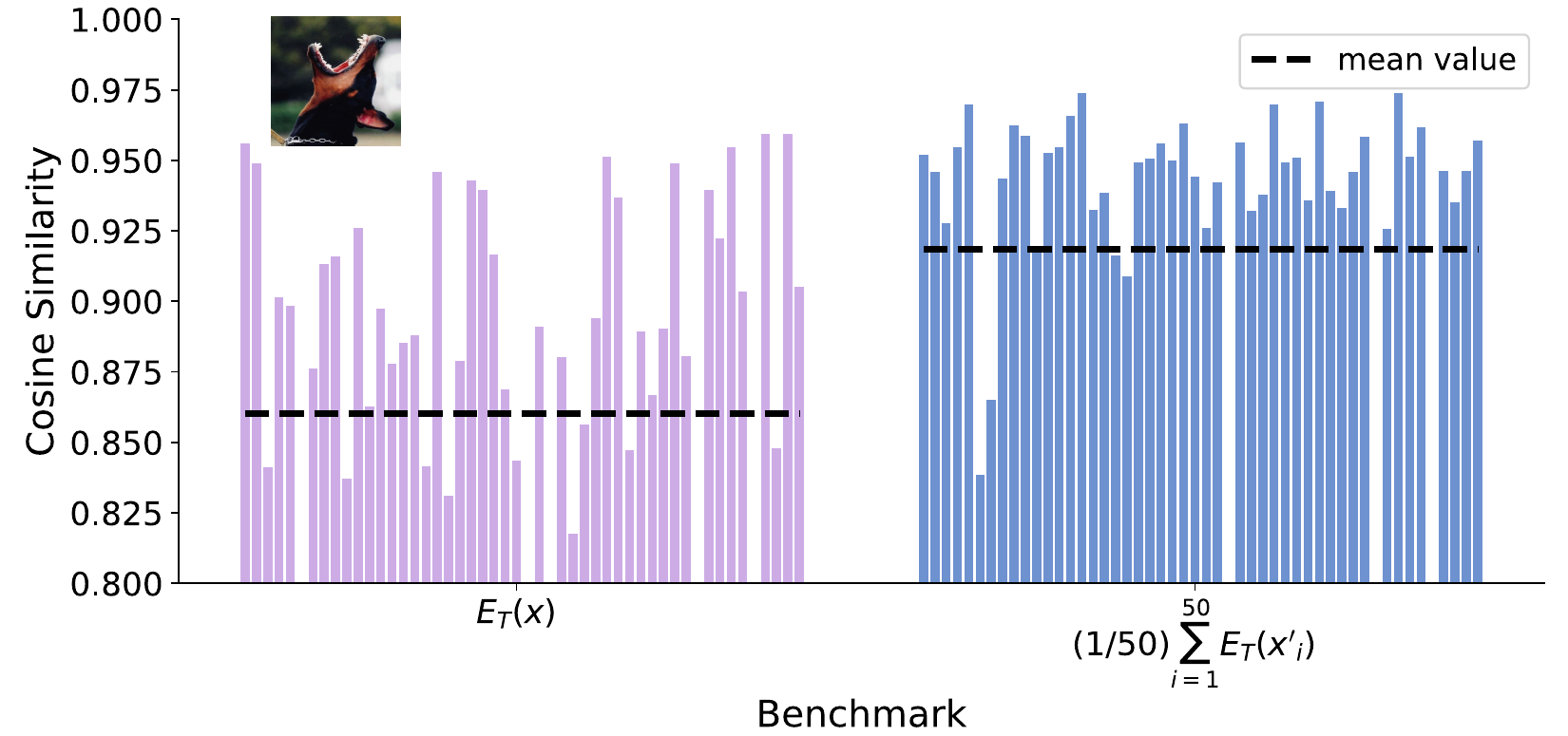}
    \caption{We augment an image $\boldsymbol{x}$ into 50 patches and feed them into an encoder $E_T$ that pre-trained on CIFAR10 (KNN Accuracy = 88.31\%, tested on CIFAR10). This figure depicts the cosine similarity between each augmentation patch's embedding and two different benchmarks, i.e., (1) the embedding of the original image and (2) the prototype of the image, with the mean value over the 50 similarities marked in the black dashed line. We can see that the prototype is significantly more similar to each patch's embedding, showcasing it is less biased.}
    \label{fig:bar_plot}
\end{figure}

\begin{figure*}[t] 
    \centering
    \includegraphics[width=\textwidth]{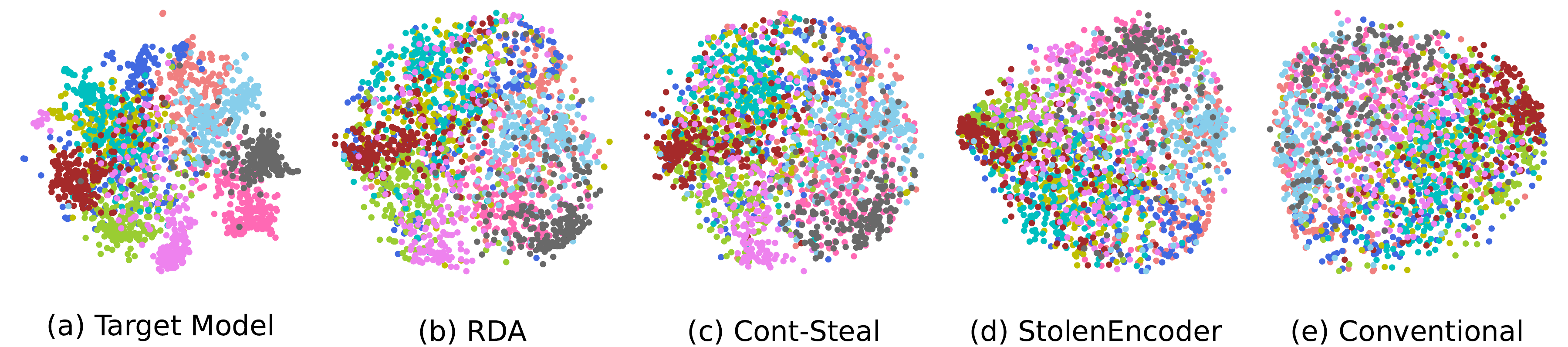}
    \caption{t-SNE of embeddings of 2,000 images randomly sampled from CIFAR10 generated by the target encoder and surrogate encoders trained with different methods.}
    \label{fig: tsne 5 methods}
\end{figure*}

\subsection{Supplementary Experiments} \label{sup: experiments}

\subsubsection{Benifit of Using Prototypes.}
We augment an image $\boldsymbol{x}$ (stamped at the upper left corner of Figure \ref{fig:bar_plot}) into 50 patches and denote each of them as $\boldsymbol{x}'_i, i=1,\dots,50$. Next, we input each patch as well as the image's original version into $E_T$, an encoder pre-trained on CIFAR10. To show the image's prototype (i.e., $\frac{1}{50}\sum_{i=1}^{50} E_T(\boldsymbol{x}'_i)$) is a less biased optimization objective compared to the embedding of its original version (i.e., $E_T(\boldsymbol{x})$), we depict the cosine similarity between each patch's embedding and the two different benchmarks in Figure \ref{fig:bar_plot}. We can see that the prototype has significantly higher similarity with each patch's embedding, showcasing it is less biased.

\subsubsection{Visualization of the Embedding Space.}
To visualize the embedding space of the surrogate encoder trained by each method, we randomly sample 200 images from each class of CIFAR10 (i.e., 2,000 images in total) and feed them to the surrogate encoder. From Figure \ref{fig: tsne 5 methods}, we can observe that embeddings of different classes from surrogate encoders trained by StolenEncoder and the conventional method overlap with each other and exhibit less structural clarity. In contrast, RDA and Cont-Steal train surrogate encoders that generate more distinguishable embeddings for different classes. This indicates that encoder stealing benefits from contrastive learning.

\subsubsection{Time Cost.}
In particular, the total training time comprises two parts, i.e., the time for training the surrogate encoder and the time for testing the trained surrogate encoder after each training epoch to choose the best one. We present the total time cost of each method in Table \ref{tab:time cost}. We can see that the testing time is 13.18 minutes and is identical across all methods over 100 epochs, while the training time of RDA is the longest. This is because RDA augments each image into 5 patches by default to train the surrogate encoder, which means 5 times forward encoding for each image. For a fair comparison, we prolong the training for another 200 epochs for each baseline method and show that our RDA still surpasses them by a large margin, as demonstrated by Figure \ref{fig: same training time}.

\begin{table}[t]
\centering
\caption{The time cost of each method over the 100 training epochs. Each result consists of two parts, i.e., the time for training the surrogate encoder and the time for testing after each epoch.}
\label{tab:time cost}
\begin{tabular}{cc}
\Xhline{1pt}
Method & Time (min) \\ \Xhline{0.65pt}
   Conventional    &   3.87 $+$ 13.18   \\ 
    StolenEncoder   &   5.53 $+$ 13.18   \\ 
    Cont-Steal   &   4.08 $+$ 13.18   \\ \Xhline{0.65pt}
    RDA   &    10.65 $+$ 13.18  \\ \Xhline{1pt}
\end{tabular}
\end{table}

\subsubsection{Effectiveness of RDA on Different Encoder Architectures.}
To demonstrate the effectiveness of our RDA against pre-trained encoders of various architectures, we further evaluate it on ResNet34, VGG19\_bn \cite{simonyan2014very}, DenseNet121 \cite{huang2017densely}, and MobileNetV2 \cite{howard2017mobilenets}. Table \ref{tab: various architectures} shows that RDA can achieve comparable performances with the target encoders of various architectures and even outperforms them on multiple datasets.

\begin{table}[t]
\centering
\caption{Results of RDA against target encoders of different architectures. The pre-training dataset is CIFAR10 and the architecture of the surrogate encoder is ResNet18.}
\label{tab: various architectures}
\resizebox{0.6\textwidth}{!}{\begin{tabular}{ccccc}
\Xhline{1pt}
\begin{tabular}[c]{@{}c@{}}Target Encoder\\ Architecture\end{tabular} & \begin{tabular}[c]{@{}c@{}}Downstream\\ Dataset\end{tabular} & TA & SA & $\frac{\text{SA}}{\text{TA}}\times 100\%$ \\ \Xhline{0.65pt}
\multirow{4}{*}{ResNet34}  & MNIST   & 96.42 & 96.19 & 99.76 \\ 
                           & F-MNIST & 89.51 & 87.03 & 97.23 \\ 
                           & GTSRB   & 62.79 & 54.57 & 86.91 \\ 
                           & SVHN    & 61.68 & 70.12 & 113.68 \\ \Xhline{0.65pt}
\multirow{4}{*}{VGG19\_bn} & MNIST   & 90.76 & 91.57 & 100.89 \\ 
                           & F-MNIST & 68.54 & 74.28 & 108.37 \\ 
                           & GTSRB   & 13.27 & 11.25 & 84.78 \\ 
                           & SVHN    & 26.32 & 41.30 & 156.91 \\ \Xhline{0.65pt}
\multirow{4}{*}{DenseNet121} & MNIST   & 95.81 & 96.15 & 100.35 \\ 
                           & F-MNIST & 86.75 & 88.79 & 102.35 \\ 
                           & GTSRB   & 54.69 & 52.05 & 95.17 \\ 
                           & SVHN    & 49.52 & 67.19 & 135.68 \\ \Xhline{0.65pt}
\multirow{4}{*}{MobileNetV2} & MNIST   & 90.56 & 94.53 & 104.38 \\ 
                           & F-MNIST & 77.69 & 83.38 & 107.32 \\ 
                           & GTSRB   & 33.88 & 24.00 & 70.84 \\ 
                           & SVHN    & 27.54 & 58.54 & 212.56 \\ \Xhline{1pt}
\end{tabular}}
\end{table}

\begin{figure*}[t]
    \centering
    \includegraphics[width=\textwidth]{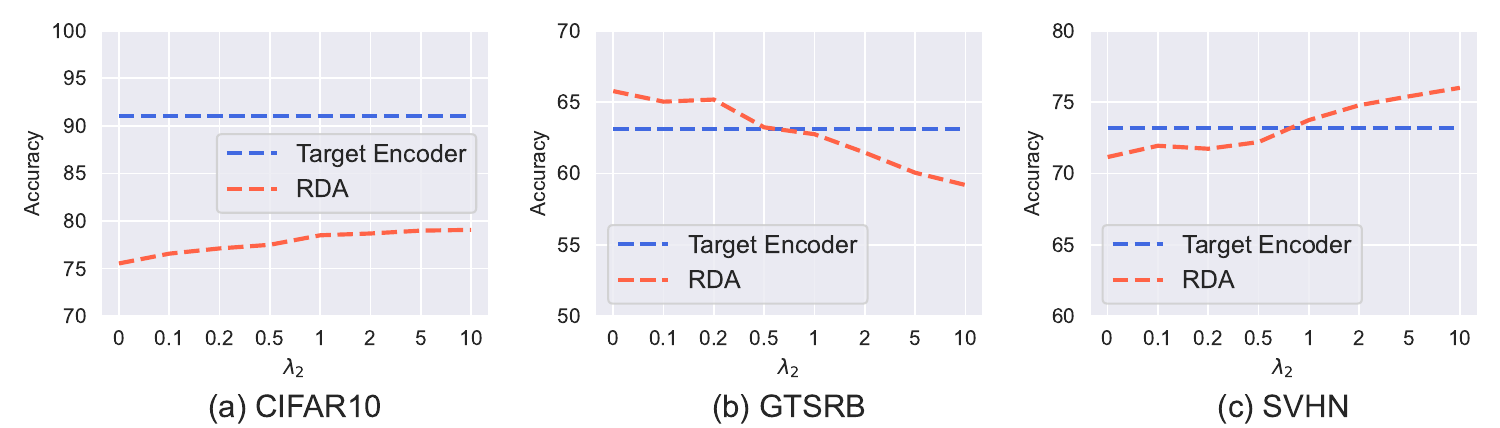}
    \caption{Ablation studies on the weight of each part in the loss, where we fix $\lambda_1$ to 1 and vary the value of $\lambda_2$ from 0 to 10.}
    \label{fig: ablations on the loss weight sup}
\end{figure*}

\subsubsection{Ablation Studies on the Weight of Each Part in the Loss Function.}
To investigate the impact of each part in our loss function on the stealing performance, we fix $\lambda_1$ in Eq. \ref{eq:8} to 1 and vary the value of $\lambda_2$ (i.e., the weight of the aligning loss $\mathcal{L}_A$) from 0 to 10. We use the encoder pre-trained on CIFAR10 as the target encoder and evaluate the trained surrogate encoder on three different downstream datasets, i.e., CIFAR10, GTSRB, and SVHN. As we can observe in Figure \ref{fig: ablations on the loss weight sup}, the surrogate encoders' performances on CIFAR10 and SVHN positively correlate to the weight of $\mathcal{L}_A$, i.e., $\lambda_2$. This is because CIFAR10 is the pre-training dataset of the target encoder. As the weight of the aligning loss $\mathcal{L}_A$ increases, the outputs of the surrogate and the target encoder become more aligned, and thus the trained surrogate encoder will perform better on CIFAR10. On the other hand, a larger weight for $\mathcal{L}_A$ will make the surrogate encoder fit the surrogate dataset more. In this sense, since SVHN is more similar to the surrogate dataset (i.e., Tiny ImageNet), the surrogate encoder naturally performs better on it as the $\lambda_2$ increases. On the contrary, as the weight of $\mathcal{L}_A$ increases, the surrogate encoder's performance on GTSRB which shares little similarity with both the pre-training and surrogate datasets becomes worse. Similar phenomenons have been discussed in Section \ref{sec: ablation studies}, where $\mathcal{L}_D$ trains the surrogate encoder that achieves the optimal performance on GTSRB. A larger weight for $\mathcal{L}_A$ will make the surrogate encoder fit the surrogate dataset more, and thus perform worse on GTSRB.

\subsubsection{Ablation Studies on the Effect of Each Part in $\mathcal{L}_A$.}
To investigate whether our $\mathcal{L}_A$ can acquire better performance compared to each part (i.e., $\mathcal{L}_{amp}$ and $\mathcal{L}_{ang}$) of it, we do ablation experiments as presented in Table \ref{tab: lamp and lang}. All settings follow the default configurations we use except the loss function. From the results, we can observe that $\mathcal{L}_D + \mathcal{L}_A$ achieves the best performance in three downstream datasets among four. Averagely, $\mathcal{L}_D + \mathcal{L}_A$ still largely surpasses the other two loss functions. This suggests the necessity of both $\mathcal{L}_{amp}$ and $\mathcal{L}_{ang}$ in our loss design of $\mathcal{L}_A$.

\begin{table}[t]
\centering
\caption{Ablations on each part of $\mathcal{L}_A$.}
\label{tab: lamp and lang}
\resizebox{0.7\textwidth}{!}{\begin{tabular}{cccccc}
\Xhline{1pt}
Loss Function & \begin{tabular}[c]{@{}c@{}} CIAFR10 \end{tabular} & GTSRB & SVHN & CIFAR100 & AVG\\ \Xhline{0.65pt}
$\mathcal{L}_D + \mathcal{L}_A$ & \colorbox[RGB]{219,216,236}{79.39} & \colorbox[RGB]{219,216,236}{62.75} & 73.74 & \colorbox[RGB]{219,216,236}{44.27} & \colorbox[RGB]{219,216,236}{65.03} \\ 
$\mathcal{L}_D + \mathcal{L}_{amp}$ & 76.17 & 56.21 & 74.40 & 40.49 & 61.81 \\ 
$\mathcal{L}_D + \mathcal{L}_{ang}$  & 77.70 & 55.85 & \colorbox[RGB]{219,216,236}{75.81} & 42.78 & 63.03\\ \Xhline{1pt}
\end{tabular}}
\end{table}

\subsubsection{Ablation Studies on the Choice of The Surrogate Data.}
To investigate the impact of the choice of the surrogate data used by attackers, we randomly sample 2,500 images from three different datasets, namely CIFAR10, GTSRB, and SVHN respectively to build three different surrogate datasets. Other settings follow the default setting. As Table\ref{tab: ablation on surrogate data} shows, using images sampled from CIFAR10 as the surrogate data offers the most effective performance. However, comparing Table \ref{tab: loss} and Table \ref{tab: ablation on surrogate data}, we can find that sampling data from CIFAR10 as the surrogate data underperformance sampling from Tiny-ImageNet on GTSRB. We suspect the reason is that GTSRB differs much from the pertaining data and thus using CIFAR10 as the surrogate data offers minimal benefits. Additionally, the diversity of Tiny-ImageNet enables it to extract broader potential representations of the target encoder.

\begin{table}[t]
\centering
\caption{Ablations on the surrogate data.}
\label{tab: ablation on surrogate data}
\resizebox{0.5\textwidth}{!}{\begin{tabular}{cccc}
\Xhline{1pt}
Surrogate Data & \begin{tabular}[c]{@{}c@{}} CIAFR10\end{tabular} & GTSRB & SVHN \\ \Xhline{0.65pt}
CIFAR10 & \colorbox[RGB]{219,216,236}{81.33} & \colorbox[RGB]{219,216,236}{59.34} & \colorbox[RGB]{219,216,236}{75.21} \\ 
GTSRB & 68.54 & 54.84 & 73.70 \\ 
SVHN  & 63.63 & 42.00 & 68.78 \\ \Xhline{1pt}
\end{tabular}}
\end{table}

\subsubsection{More Complex Downstream Datasets.}
We have also conducted small-scale experiments on three more complex downstream datasets: Tiny-ImageNet, CUB-200-2011, and Food 101. For these datasets, we center-crop and resize each image to $224 \times 224$, followed by training the downstream classifier for 300 epochs to attain relatively decent performances. The results presented in Table \ref{tab: more complex datasets} demonstrate that our RDA consistently delivers superior performances.

\begin{table}[t]
\centering
\caption{Additional experiments on more difficult datasets. The reported result is \textbf{$\frac{\text{SA}}{\text{TA}}\times 100\%$}, which shows the attack efficacy.}
\label{tab: more complex datasets}
\resizebox{0.8\textwidth}{!}{\begin{tabular}{cccccc}
\Xhline{1pt}
 Dataset & Description & Conventional & StolenEncoder & Cont-Steal & RDA \\ \Xhline{0.65pt}
 Tiny-ImageNet & 200 classes & 34.29 &  79.81  & 92.14 &  \colorbox[RGB]{219,216,236}{94.76}  \\
 CUB-200-2011 & 200 classes & 31.35 &    44.58 & 54.88 &  \colorbox[RGB]{219,216,236}{57.37}\\
 Food101 & 101 classes & 28.77 &  40.04  & 79.29 & \colorbox[RGB]{219,216,236}{79.77} \\ \Xhline{1pt}
\end{tabular}}
\end{table}

\subsubsection{Stealing Third-Party Models.}
To mitigate the potential risks of violating relevant laws, we refrain from unauthorized usage of commercial APIs for stealing attacks. Consequently, to further validate the applicability of our method, we provide additional results using open-sourced third-party models in accordance with our default settings for a simulation. After deploying the target model, we assume knowledge only of the dimension of the output embeddings. Results are presented in Table \ref{tab: third-party model}, illustrating the applicability of RDA on these third-party models.

\begin{table}[t]
\centering
\caption{Attack efficacy (\textbf{$\frac{\text{SA}}{\text{TA}}\times 100\%$}) against third-party models. \textbf{These target models can be downloaded via our code link.}}
\label{tab: third-party model}
\resizebox{0.8\textwidth}{!}{\begin{tabular}{cccc}
\Xhline{1pt}
Provider / Platform & URL & CIFAR10 & CIFAR100 \\ \Xhline{0.65pt}
OpenMMLab & https://github.com/open-mmlab &  84.72 &  107.59  \\
Hugging Face & https://huggingface.co & 99.53 &  91.41  \\ \Xhline{1pt}
\end{tabular}}
\end{table}

\subsection{Possible Defenses against RDA} \label{sup: possible defense}
Since our RDA augments each image into multiple semantically similar patches, this may result in frequent similar queries at the embedding level. Therefore, a potential defense against RDA could involve rejecting such frequent and similar queries. For instance, if a user's recent queries exhibit high similarity, the defender might reject subsequent queries. However, this approach has several challenges. Determining the appropriate threshold for rejecting similar queries is complex and requires careful consideration. Moreover, the defender might need to dynamically adjust the threshold and rejection criteria. Additionally, if the user distributes the queries across multiple accounts, the defense becomes even more challenging. Effective defense strategies still require significant development.

\end{document}